\documentclass{article}
\PassOptionsToPackage{numbers, compress}{natbib}
\usepackage[final]{neurips_2023}

\usepackage[utf8]{inputenc} % allow utf-8 input
\usepackage[T1]{fontenc}    % use 8-bit T1 fonts
\usepackage{hyperref}       % hyperlinks
\usepackage{url}            % simple URL typesetting
\usepackage{booktabs}       % professional-quality tables
\usepackage{amsfonts}       % blackboard math symbols
\usepackage{nicefrac}       % compact symbols for 1/2, etc.
\usepackage{microtype}      % microtypography
\usepackage{xcolor}         % colors

\usepackage{graphicx}       % 이미지
\usepackage{kotex}          % 한글
\usepackage{wrapfig}        % 중간에 figure 넣기
\usepackage{adjustbox}
\usepackage{array,multirow} % multirow
\usepackage{tcolorbox}

\usepackage{subcaption}   
\usepackage{amsmath}
\usepackage{amssymb}
\usepackage{array}
\usepackage{times}
\usepackage{epsfig}

\title{Semantic Image Synthesis with \protect\\ Unconditional Generator }
\author{%
  Jungwoo Chae$^{12}$\thanks{Equal contribution.}\ \ \ 
  Hyunin Cho$^{1}$\footnotemark[1]\ \ \
  Sooyeon Go$^{1}$\ \ \
  Kyungmook Choi$^{1}$ \ \ \
  Youngjung Uh$^{1}$\thanks{Corresponding author}\\
  \\
    $^{1}$Yonsei Unviersity, Seoul, South Korea\\
    $^{2}$LG CNS AI Research, Seoul, South Korea\\
  \tt\small\texttt{\{cjwyonsei, hyunin9528, sooyeon8658, kyungmook.choi, yj.uh\}@yonsei.ac.kr}\\
  \tt\small\texttt{\{cjwoolgcns\}@lgcns.com}
}

\newcommand{\fref}[1]{Figure \ref{#1}}

\newcommand{\tref}[1]{Table \ref{#1}}
\newcommand{\aref}[1]{appendix}

\newcommand{\uh}[1]{\textcolor{red}{}}

\newcommand{\cj}[1]{\textcolor{blue}{}}

\newcommand{\ch}[1]{\textcolor{blue}{}}

\newcommand{\ck}[1]{\textcolor{green}{}}

\newcommand{\gs}[1]{\textcolor{green}{}}

\def\vm{{\mathbf{m}}}

\def\vf{{\mathbf{f}}}

\def\vz{{\mathbf{z}}}

\begin{document}

\maketitle

\vspace{-4mm}

\begin{abstract}

Semantic image synthesis (SIS) aims to generate realistic images that match given semantic masks. Despite recent advances allowing high-quality results and precise spatial control, they require a massive semantic segmentation dataset for training the models. Instead, we propose to employ a pre-trained unconditional generator and rearrange its feature maps according to proxy masks. The proxy masks are prepared from the feature maps of random samples in the generator by simple clustering. The feature rearranger learns to rearrange original feature maps to match the shape of the proxy masks that are either from the original sample itself or from random samples.
% As we do not rely on manual annotations, we design a feature rearranger that rearranges original feature maps according to proxy masks. The proxy masks are easily obtained by clustering feature maps of random samples.
Then we introduce a semantic mapper that produces the proxy masks from various input conditions including semantic masks.
% Our intuition lies in that the feature maps already have enough semantics thus the feature maps should be transformed if the semantics change. We call this semantics a proxy mask for generation and design a semantic mapper that translates various input conditions including semantic segmentation maps.
Our method is versatile across various applications such as free-form spatial editing of real images, sketch-to-photo, and even scribble-to-photo. Experiments validate advantages of our method on a range of datasets: human faces, animal faces, and buildings. 

\end{abstract}
\vspace{-3mm}

\begin{figure}[h]
  \centering
  \includegraphics[width=\linewidth]{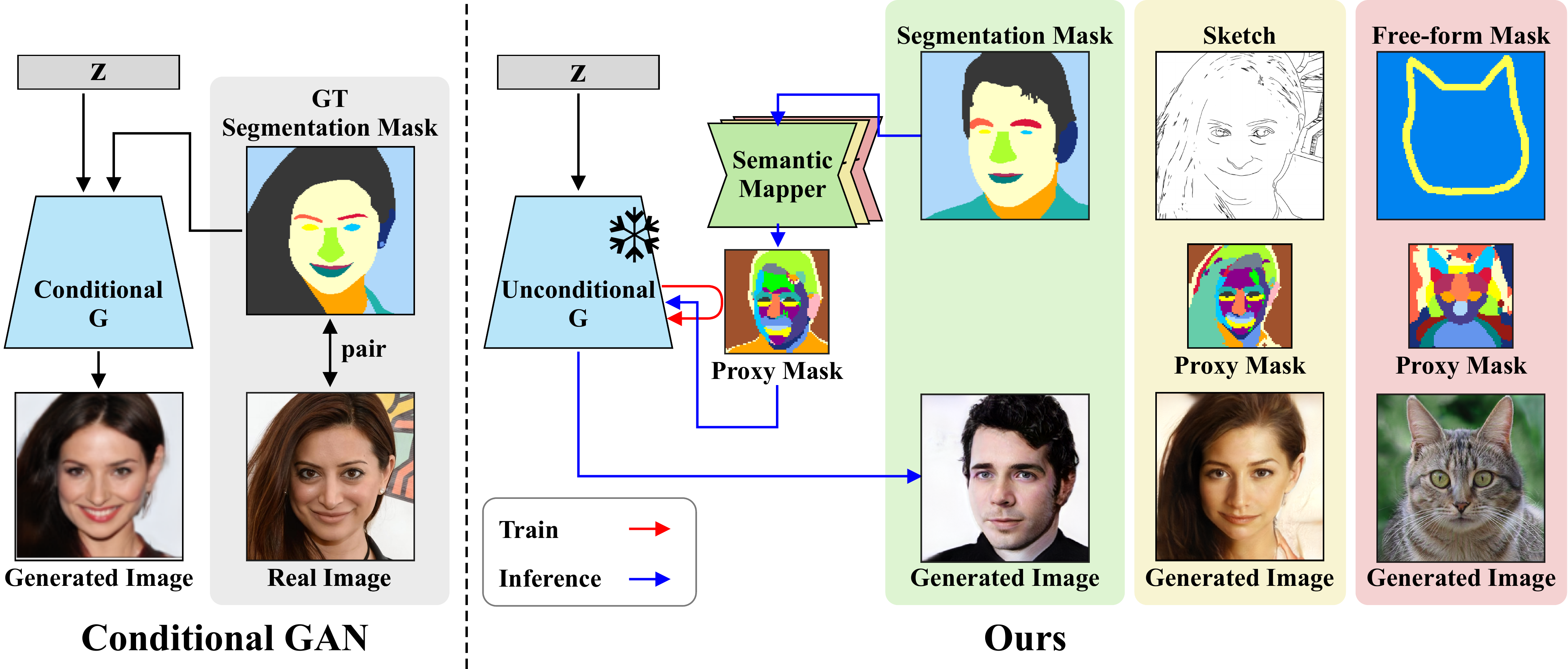}
  \caption{\textbf{Conditional GANs vs Ours} (Left) Conditional GANs train a generator using expensive pairs of semantic masks and images.
  % from scratch by utilizing multiple sets of prepared image-mask pairs.
  (Right) On the other hand, our method does not require a large number of image-mask pairs in training process and uses a pre-trained unconditional generator for semantic image synthesis. Furthermore, it accommodates various types of inputs such as sketches or even simpler scribbles. 
  % We have the flexibility to accept various types of input masks, such as segmentation masks, sketches, and free-form masks, as conditions.
  }
  \label{fig:teaser}
  \vspace{-3mm}
\end{figure}

\section{Introduction}

Semantic image synthesis (SIS) aims to synthesize images according to given semantic masks. 
Its ability to reflect the spatial arrangement of the semantic masks makes it a powerful tool for various applications such as scene generation\cite{spade, oasis} and photorealistic image editing\cite{sdm}.
However, it requires abundant pairs of images and per-pixel class annotations which are laborious and error-prone. If we want to change the granularity of control, we need to re-annotate the images and re-train the model. E.g., a network pretrained on human faces and their semantic segmentation masks cannot be used for generating human faces from simple scribbles and requires re-training on scribble annotation.

% 문장 연결을 부드럽게 고쳤습니다: SIS의 정의를 먼저 제시 -> 뭐가 강점이다 -> 그래서 뭐에 유용하다
% It is a promising approach for a variety of applications such as photorealistic image editing, virtual try-on, and scene generation. Its ability to facilitate precise control of image generation by leveraging segmentation masks makes it a powerful tool for these applications. 
% 가독성 향상: 문장 구조 단순화 및 분리
% However, the need for laborious and error-prone per-pixel class annotations, especially when the number of annotation classes changes, hampers the scalability of existing SIS methodologies.
% 아래 문장은 기존의 단점에서 언급하기보다는 우리의 장점으로 언급하는게 더 유리하겠습니다. class 수를 줄이는건 simple preprocessing으로 가능해서 별로 단점같지 않군요.
%Furthermore, if we change the segmentation configuration, e.g., foreground mask instead of 7-class cat segmentation mask (\fref{fig:teaser} right), we should re-train a model for the new configuration.

% rely heavily on the learned distribution of off-the-shelf models such as semantic segmentation network, and 
A few studies \cite{panda, edit} have attempted to bypass this issue by employing pretrained unconditional generators for image manipulation without the need for annotations. These approaches, however, struggle to allow detailed manipulation, presenting challenges for users desiring to customize generated images.
Furthermore, users have to manually identify relevant clusters and tweak feature maps, leading to burdensome manipulation process.
% which can result in a decrease in image quality when deforming the feature maps.

%While these methods lessen the reliance on annotated data, they frequently fall short in providing detailed manipulation, posing challenges for users desiring to customize the generated images. Users must manually identify corresponding clusters and tweak feature maps, a process often leading to an undesirable decline in image quality during the deformation of the feature maps.

% An example of this limitation is "Editing in Style\cite{edit}", which deforms feature maps using a pretrained GAN to edit images, but struggles with editing unaligned images. %Furthermore, these methods are not capable of generating images outside of certain distributions, such as the CLIP\cite{clip}.

In response to these challenges, we introduce a novel method for semantic image synthesis: rearranging feature maps of a pretrained \textit{unconditional} generator according to given semantic masks.
% We employ a pretrained unconditional generative adversarial networks (GANs) and rearrange its feature maps based on an input semantic mask.
We break down our problem into two sub-problems to accomplish our objectives: rearranging and preparing supervision for the rearrangement. First, 
% 장황하고 반복돼서 뺐어요 
% we rearrange feature maps to align with the layout of the input semantic mask. Second, we provide pairs of input mask and rearranged feature maps solve the lack of paired inputs and outputs for the first problem. 
we design a rearranger that produces feature maps rearranged according to inputs by aggregating features through attention mechanism. 
% We introduce a novel rearranging network to handle these challenges, which naturally reorders feature maps in a way that mirrors the input semantic mask.
Second, we prepare randomly generated samples, their feature maps, and spatial clustering. Note that the samples are produced online during training thus it does not require additional storage.
% To address the data shortage for training the rearranging network, we propose a self-supervised approach that capitalizes on the representation of pretrained generator.
% In combined, we design
This approach assumes that we have two random samples and their intermediate feature maps from the generator. The rearranger learns to rearrange the feature maps of one sample to match the semantic spatial arrangement of the other sample. 
% 이하 중복 reconstruct를 살리고 싶으시면 match 자리에 쓰세요
%, by reconstructing the arrangement of it. 

% feature map을 input mask를 받아 바꿔야 하고
% 이때 두 가지의 subproblem이 있는데 한 가지가 semantic mask를 맞추어 자연스럽게 arrange 하는것, 다른 한 가지가 annoatation 문제를 해결하면서 학습하는 것
% Rearragne network를 통해 첫번째를 해결 두번째는 proxy mask 개념을 말하면서 input output pair를 만들었다는걸 설명
% As we do not assume a large-scale semantic segmentation dataset, we first condition the rearranger on a proxy mask which is rough semantics of the feature maps.
% 언컨디셔널 제너레이터가 자기자신을 슈퍼바이스하게 한다.
% input과 target이 궁금해질텐데 그때 proxy를 소개하는게 어떨까
% The self-guided proxy mask eliminates the need for additional annotation. And the rearranging process allows for precise image editing that reflects the mask's semantics, regardless of the model or training data distribution. To facilitate this rearrangement of feature maps, we introduce a rearranging network, which is trained using pairs of generator feature maps and their matching generated images, thus reducing the burden of annotation.

% However, when an input mask is received, a semantic gap occurs between the user input and the proxy mask from the generator. As shown in \fref{fig:teaser}, there is a discrepancy between the proxy mask that the generator can comprehend and the input mask provided by humans.

However, a semantic gap may arise when the generator receives an input mask. As shown in \fref{fig:teaser}(right), this gap is manifested in the discrepancy between the input segmentation mask and the proxy mask. The proxy masks are the arrangement of feature maps which are derived from the feature maps in the generator through simple clustering. We solve this perception difference by training a semantic mapper that can convert an input mask to the proxy mask. 

%\todo{flexibility와 various application 정리하기}
%This semantic mapper is capable of accepting various types of input conditions instead of an input mask such as scribbles, Holistically-Nested Edges(HED)\cite{hed}, and Canny Edges, allowing semantic image synthesis.

%Furthermore, if we change the segmentation configuration, e.g., foreground mask instead of 7-class cat segmentation mask (\fref{fig:teaser} right), we should re-train a model for the new configuration.

Our proposed method offers several key advantages. Not only does it significantly reduce the training time and the burden of manual annotation, but it also leverages pretrained GANs to yield high-quality synthesized images. Additionally, once we train the rearranger, the need to only train the semantic mapper, even when the segmentation configuration changes, makes our approach computationally efficient compared to methods requiring retraining of the entire model. Moreover, it is compatible with other latent editing techniques such as StyleCLIP\cite{styleclip}.

As our method operates through feature maps rearrangement, it enables free-form mask spatial editing with fewer constraints on the mask's shape. Furthermore, our semantic mapper accepts different types of input conditions such as HED~\cite{hed} and Canny Edges. Therefore, our approach can be applied to various applications like sketch-to-photo and scribble-to-photo transformations.

We demonstrate pixel-level content creation with a pretrained generator on several datasets including CelebAMask-HQ, LSUN Church, and LSUN Bedroom. We also present real image editing examples where the source image is edited to conform to the shape of the target image or target mask. Lastly, our method achieves noticeably higher mIOU scores and significantly lower FID than prior works. Furthermore, our method outperforms the baselines in both qualitative and quantitative measures. 

%% chat gpt said
% 1. Annotation Efficiency
% 2. Training Efficiency
% 3. Flexibility
% 4. Precise Control
% 5. Scalability
% 6. Compatibility
% 7. Quality of Synthesized Images
% 8. Perceptual Alignment
% 9. Superior Performance
\section{Related Work} \label{sec:related_work}

\paragraph{Semantic image synthesis}
Semantic Image Synthesis (SIS) is a specialized type of conditional image generation that can accurately mirror user intentions at pixel level~\cite{pix2pix, pix2pixhd, dsgan, ccfpse, dagan, scgan, spade, sean, oasis, inade}. Over time, the field has advanced significantly, improving controllability and thus enabling the production of diverse, high-quality results. These advancements also allow for the adjustment of the style of individual class components based on semantic masks. Like preceding GAN methods, the application of diffusion in SIS also facilitates the alteration of image style under specific conditions~\cite{sdm}.  More recently, diffusion-based conditional image generation has taken this further, creating high-quality images using semantic masks and other inputs like edge maps and user scribbles~\cite{ddpm, sketchguidediffusion, piti, ldm, controlnet}. Despite their benefits, SIS methods are resource-intensive, requiring many image-label pairs during learning. Moreover, any changes in annotations necessitate a complete retraining of the model, leading to inefficiencies. Unlike these methods, our approach uses pairs of generator feature maps and the corresponding generated images as data, effectively eliminating the need for expensive image-label pair datasets. As a result, our method provides significant cost and time savings compared to previous methods.

\paragraph{Feature map manipulation for semantic part editing}
Manipulating feature maps of a generator directly influences the resulting synthesized image, and the clustering results of these feature maps yield semantic part representations~\cite{edit}. Prior works have leveraged these characteristics of feature maps to conduct local editing tasks such as enlarging the eyes or changing the shape of ears without manual annotations \cite{panda}. 
However, these methods are inconvenient when it reflect the intention of user during the image editing process, presenting a considerable challenge. 
To address this, we propose a novel approach that transforms the feature maps into the shape of a given mask, allowing more intuitive and precise image generation. This approach significantly enhances the user experience by providing an efficient and convenient means of generating images that closely align with the user's intention.

\paragraph{Mapping between feature maps and semantics}%Feature maps as a pixel-level annotation}
The utility of feature maps extends beyond image generation to tasks such as few-shot semantic segmentation, offering rich semantic information from a pretrained generator\cite{semanticgan, datasetgan, lineargan, repurposegan, sgone, ddpmseg}. Techniques such as linear transformation or a simple CNN network can generate high-quality annotations from a minimal set of label-image pairs\cite{lineargan, repurposegan}. While the studies mentioned above utilize feature maps to generate segmentation maps, our research goal takes a reverse approach. Inspired by \cite{edit}, ours create proxy masks based on the information provided by segmentation maps or other conditions so that the model can rearrange feature maps based on the transferred clusters. Notably, LinearGAN\cite{lineargan} excels in enabling few-shot segmentation and achieving semantic image synthesis by finding the optimized latent codes.
% mask에 가장 적합한 latent code를 찾는다는 의미 삽입.
However, optimization in the latent space in LinearGAN lacks detailed pixel-level control and makes it challenging to apply user-desired styles properly. Our method conducts semantic image synthesis at the feature map level, significantly enhancing pixel-level control and allowing users to apply various styles from references in their preferred manner.

\paragraph{Exemplar guided image synthesis}
Utilizing exemplar images rather than varying domain conditions could provide a broader controllability range. Most exemplar image generation problems aim to stylize the given condition with the style of the exemplar image \cite{cocosnet, cocosnet2, dynast, matebit}. However, we focus more on reflecting the shape of the exemplar image. Techniques such as SNI\cite{sni}, DAT\cite{diagonalgan}, TransEditor\cite{transeditor}, and CoordGAN\cite{coordgan} use dual latent spaces for image editing, which helps to disentangle the structure and style of an image. By incorporating an encoder to extract the structural latent codes of the exemplar image, it is possible to generate images that retain the shape of the exemplar while exhibiting a range of stylistic variations. However, these approaches predominantly rely on latent codes, making it challenging to replicate the structure of exemplar images precisely. While similarly employing latent codes, our method enhances spatial manipulation accuracy by rearranging the feature maps. 

\paragraph{Foundation model for segmentation}
The Segment Anything Model (SAM)\cite{sam} is a foundational vision model trained on over 1 billion masks. With SAM, zero-shot segmentation for general images becomes possible. The method allows the creation of segmentation masks for entire image datasets, which can then be used to train conventional SIS models. However, SAM does not cover every domain-specific semantics. Additionally, utilizing previous SIS models with SAM still requires the time-consuming process of generating mask pairs and training SIS models from scratch. Our method addresses situations where foundational models like SAM may not be suitable. It can be applied to user-specific custom datasets, offering a more efficient and adaptable solution.

% \section{Self-supervised Semantic Image Synthesis Using StyleGAN}
\section{Method}
% \cjw{현재 제목 vs Method}

% 아예 self랑 cross랑 나눠서 보여주기

In this section, we explain our problem formulation and its components.
\fref{fig:teaser} compares the conventional semantic image synthesis and our method. The conventional semantic image synthesis methods require pixel-wise semantic label maps for all images. Conversely, our goal is to provide spatial control of a pretrained unconditional generator using an input mask. We formulate the problem as rearranging intermediate features in the generator according to the input mask. As the means for providing spatial control, we introduce a proxy mask, which is defined by unsupervised clustering of the features. Then, we train a rearranger for rearranging the features according to the proxy mask. We introduced a semantic mapper that maps the input condition to the proxy mask to bridge the perception gap between the proxy mask and the input mask.

\subsection{Self-supervised learning of spatial control} \label{sec:proxy}

Our primary objective is to spatially rearrange the feature maps in the unconditional generator to generate appropriate images according to the specific conditions. To this end, we introduce a proxy mask that specifies the target arrangement and a rearranger, a network that utilizes an attention mechanism. Since attention is a weighted sum of values based on the similarity of the query and the key, it has proven effective for condition-based input modifications to rearrange the feature maps according to the given proxy mask.

\begin{figure}[t!]
  \centering
  \begin{subfigure}[b]{0.49\textwidth}
    \centering
    \includegraphics[width=\textwidth]{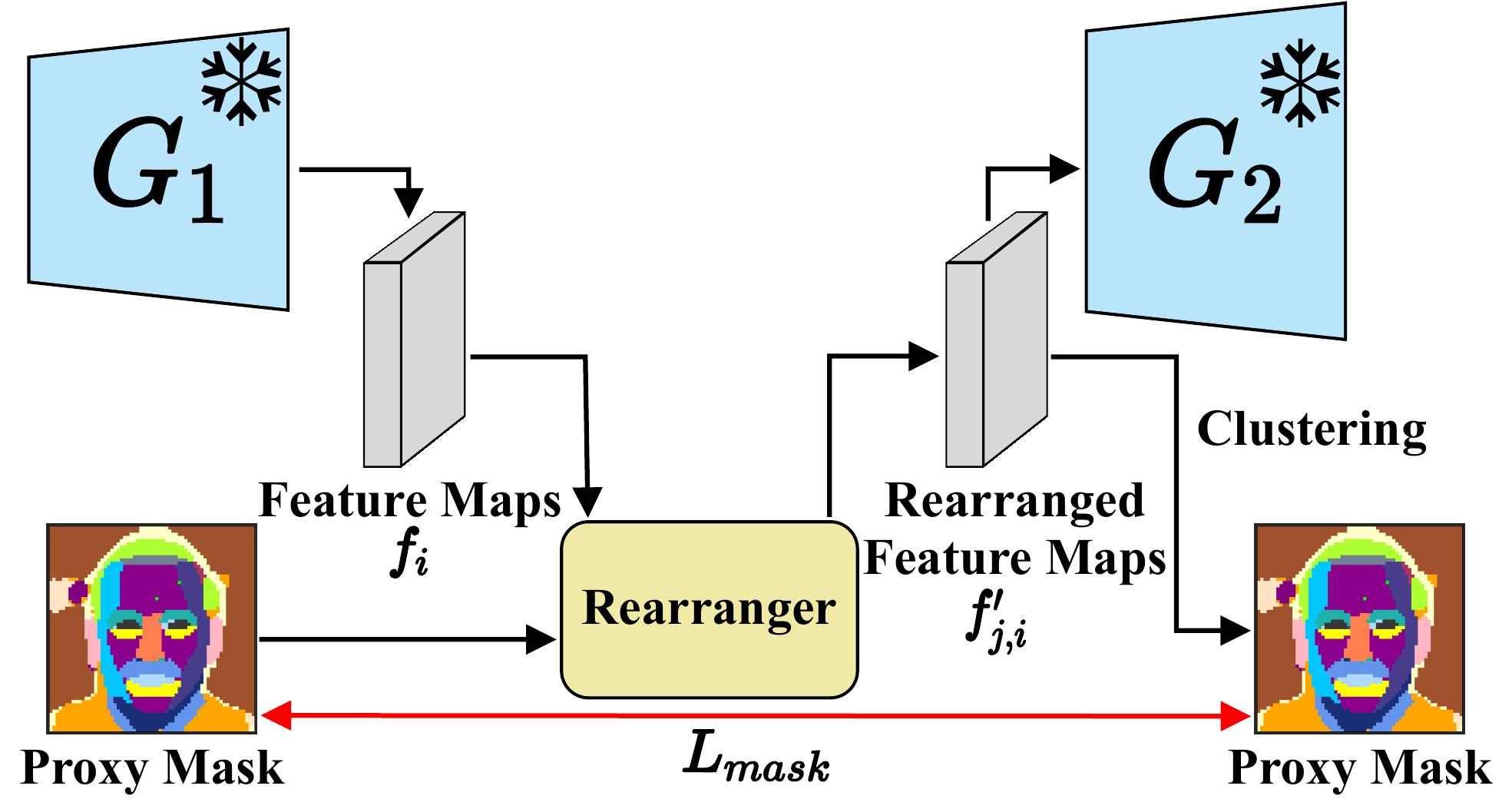}
    \caption{\textbf{Mask Loss}}
    \label{fig:mask}
  \end{subfigure}
  \hspace{0.01\textwidth} 
  \vline
  \hspace{0.01\textwidth} 
  \begin{subfigure}[b]{0.47\textwidth}
    \centering
    \includegraphics[width=\textwidth]{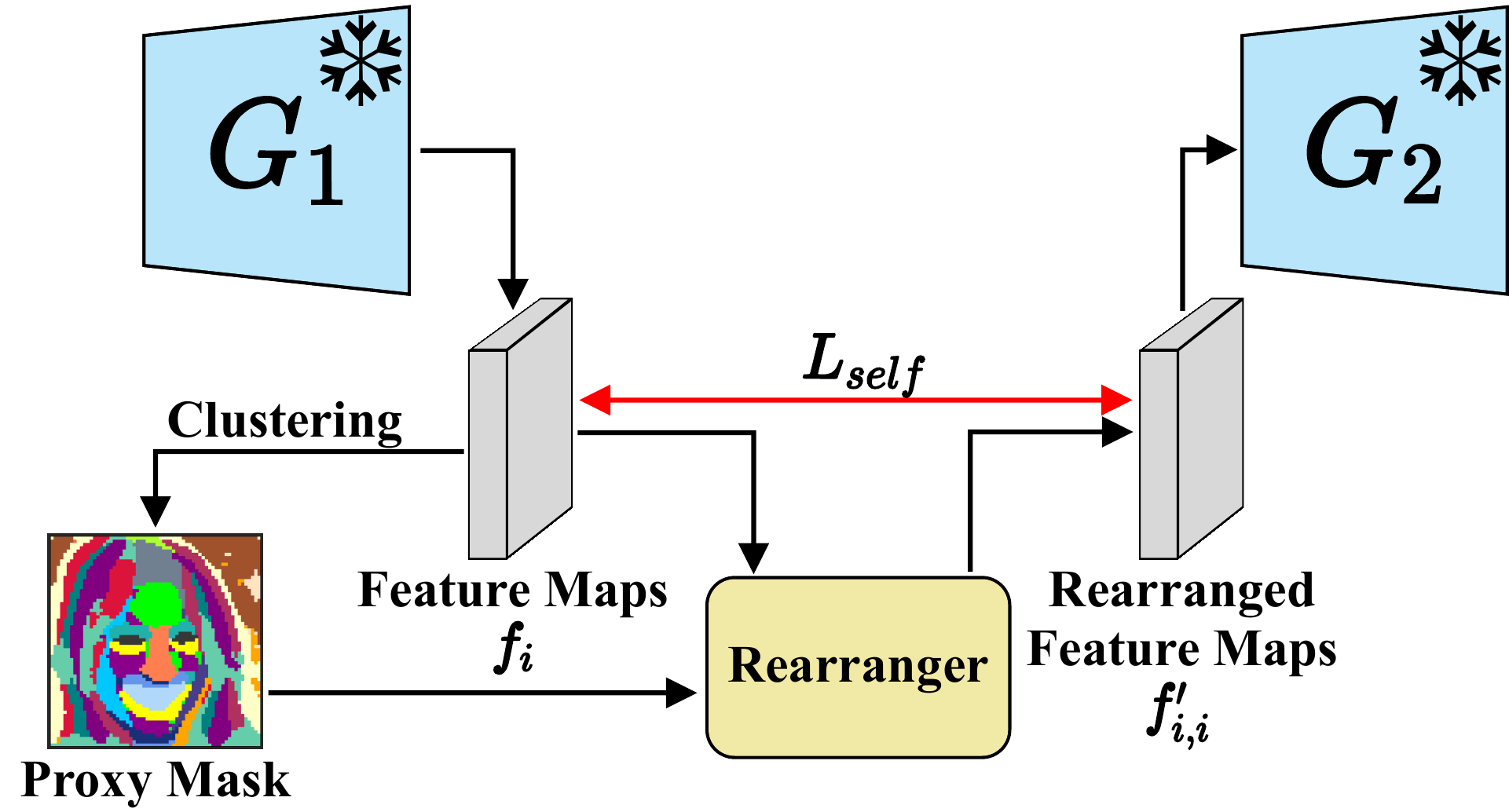}
    \caption{\textbf{Self-reconstruction Loss}}
    \label{fig:recon}
  \end{subfigure}
  \caption{\textbf{Rearranger} achieves self-supervised learning by incorporating self-reconstruction loss and mask loss, reducing the reliance on a large image-mask dataset. $G_1$ is earlier layers of the generator and $G_2$ is later layers after the proxy mask resolution.}
  \label{fig:spatial_control}
\end{figure}

% m_proxy denotes proxy mask, $K$ denotes K-means clustering algorithm,

Given a proxy mask $\vm_i$ and original feature maps $\vf_i=G_1(\vz_i)$ where $\vz$ is a random latent code, i denotes an index of the sample and $G_1$ is the front part of the model up to and including the layer corresponding to the feature maps used to generate the proxy mask, a rearranger produces feature maps through a cross-attention mechanism: $\text{Attention}(Q,K,V)= \text{softmax}\left(\frac {Q K^T}{\sqrt {d}}\right) V$. Here, $Q$ denotes the embedding of the proxy mask $\vm_i$ which is a non-linear transformation of the proxy mask., $K$ and $V$ denotes the embedding of the original feature maps $\vf_i$, $d$ denotes the dimension of the embedding space. Formally,
\begin{equation}
Q = W_Q\cdot\vm_i,\quad K = W_K\cdot\vf_j, \quad V = W_V\cdot\vf_j,
\end{equation}
where $W_{(\cdot)}\in\mathbb{R}^{d\times d_\vf}$ are learnable embedding matrices~\cite{perceiver, attention}. For simplicity, we use rearranged feature maps as $\mathbf{f}^\prime_{i,j} = \text{Attention}(\vm_i,\vf_j,\vf_j)$ instead of $\text{Attention}(Q,K,V)$. With this attention mechanism, the image rendering can be represented as $\mathbf{X}_\text{fake}=G_2(\mathbf{f}^\prime_{i,j})$, where ($i$, $j$) is a random pair of latent codes, $\mathbf{X}_\text{fake}$ denotes the synthesized image and $G_2$ is the rest part of generator consists of later layers of generator. Further, we define proxy mask $\vm_i$ in terms of $\vf_i$ via $\text{Clustering}$ which is a K-means clustering algorithm , expressed as $\vm_i=\text{Clustering}(\vf_i)$.

As shown in \fref{fig:spatial_control}, to make output of rearranger more accurately reflect the mask given as condition, we encourage the network to reconstruct the proxy mask from the rearranged feature maps by 
\vspace{-0mm}
\begin{align} 
& \mathcal{L}_{\operatorname{mask}} = 
     \text{CrossEntropy}(\mathbf{m}^\prime_{i,j},\mathbf{m}_{i})
\end{align}
Where $\mathbf{m}^\prime_{i,j} = \text{Clustering}(\mathbf{f}^\prime_{i,j})$ denotes the proxy mask of rearranged feature maps.

% To make Rearranger's output more accurately reflect the mask entered as a condition, use the

% when the proxy mask is created from a different latent, the rearranger may tend to ignore the proxy mask, which leads to the original feature map being produced. To prevent this,

% \begin{equation}
%     \mathcal{L}_{\text{mask}} = 
%      \left\Vert\fK(\text{Attention}(\fK(\vfo),\vfo,\vfo)) - \fK(\vfo)\right\Vert_2.
%     \label{eq:maskrecon}
% \end{equation}

% \vspace{-5mm}
% \begin{equation}
%     \mathcal{L}_{\operatorname{mask}} = 
%      \text{CrossEntropy}(\mathbf{m}^\prime_{i,j},\mathbf{m}_{i}),
%      \label{eq:maskrecon}
% \end{equation}

However, to maintain the style of source latent, we encourage the network to self-reconstruct when a proxy mask created from the same latent, is used as the query for attention:

\vspace{-0mm}
\begin{align} 
    & \mathcal{L}_{\text{self}} =  \sum_{i} {\left\Vert \mathbf{f}^\prime_\text{i,i} - \vf_i \right\Vert}_2
    \label{eqn:loss self}
\end{align} 

% However, it's easy to lose the style of the source in the learning process.

% \begin{equation} 
%     \mathcal{L}_{\text{self}} = 
%      \sum_{l} {\left\Vert \mathbf{f}^\prime_\text{i,i} - \vf_i \right\Vert}_2.
%     \label{eq:self}
% \end{equation}

% Furthermore, we synthesize resulting images by rendering the output of rearranger $G_2(\mathbf{f}^\prime_{i,j})$, where ($i$, $j$) is a random pair of latent codes.

Our overall learning pipeline operates as follows. We first prepare the original feature maps $\vf_{i}$ from a random latent code $\vz_{i}$. We then apply K-means clustering on the feature maps to derive segmentation-like masks. We apply the loss as mentioned above functions using these masks and feature maps, which are created from different latents. These processes work in a self-supervised manner, enabling the rearranger to learn effectively and autonomously by leveraging the representation of the pretrained generator and an unannotated dataset.

Finally, the overall training term, denoted as $\mathcal{L}_{\operatorname{total}}$, is composed of the non-saturating adversarial loss with R1-regularization~\cite{goodfellow, mescheder}, self-reconstruction loss, and mask reconstruction loss.

\begin{align} 
    & \mathcal{L}_{\text{total}} = \mathcal{L}_{\text{adv}} + \lambda_{\text{self}}\mathcal{L}_{\text{self}} + \lambda_{\text{mask}}\mathcal{L}_{\text{mask}} + \lambda_{\text{R1}}\mathcal{L}_{\text{R1}} 
    \label{eqn:total}
\end{align}

%s\vspace{10mm}
\subsection{Semantic guide} \label{sec:gap}

\begin{wrapfigure}{r}{0.5\textwidth}
  \vspace{-5mm}
  \includegraphics[width=0.5\textwidth]{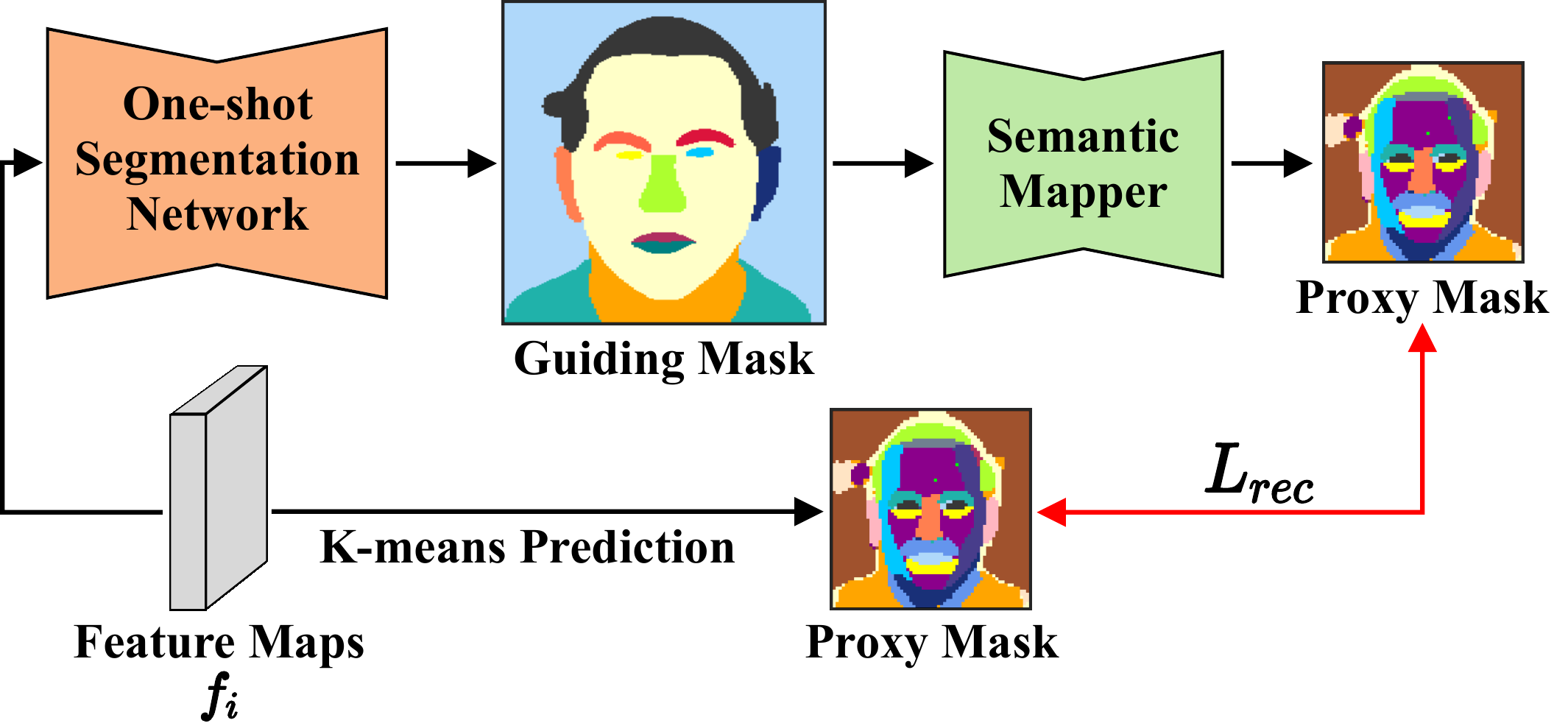}
  \caption{\textbf{Semantic Mapper} transforms input masks that the generator cannot comprehend into proxy masks, enabling the generator to understand and synthesize the corresponding output. When training the semantic mapper, a one-shot segmentation network and reconstruction loss are used.}
  \label{fig:mapper}
\end{wrapfigure}

In the earlier sections, we have introduced an approach to provide spatial control for a pretrained unconditional GAN through proxy masks generated via unsupervised feature maps clustering. However, the process lacks a direct connection from the input mask to the proxy mask. To bridge this gap, we propose a semantic mapper to convert the user given condition including input mask into a proxy mask, as illustrated in \fref{fig:mapper}. This conversion can be represented as $\text{Mapper}(\mathbf{c})$, where $\mathbf{c}$ is the input condition.

We begin the process by generating images and their corresponding proxy masks from random latent codes. Next, we employ a one-shot segmentation network, as suggested in \cite{repurposegan}, to prepare a set of mask and proxy mask pairs. These pairs are used to train our semantic mapper, allowing it to generate a proxy mask $\vm_i$ from the input mask $\vm^\prime_i= \text{SegNet}(\bar{\vf_i})$ by 
\vspace{-0mm}
\begin{align} 
    & \mathcal{L}_{\text{rec}} = \text{CrossEntropy}(\vm^\prime_i, \vm_i)
    \label{eqn:loss recon}
\end{align}
 Here, SegNet denotes the one-shot segmentation network and $\bar{\vf_i}$ is the concatenation of several feature maps for the SegNet input. Through this training, we establish a connection that bridges the perception gap between human and the generator, enabling the generator to better understand and integrate the semantic content intended by the user.

% \begin{align} 
%         & \mathcal{L}_{\operatorname{rec}} = 
%          \sum_{i} {\left\Vert \mathbf{G}^{l}(z_i, \mathbf{K}(\mathbf{G}^{l}(z_{i})) - \mathbf{G}^{l}(z_i) \right\Vert}_2
%         \label{eqn:Rec}
% \end{align}

% \vspace{-5mm}
% \begin{align}
% & \mathcal{L}_{\text{rec}} =
%  \text{CrossEntropy}(\mathbf{M}(\vm^\prime_i, \vm_i))
% \label{eqn:Rec}
% \end{align}

Furthermore, we have also incorporated an adversarial training strategy in the semantic mapper's training process. We utilize a discriminator identical to the one used in StyleGAN2 \cite{stylegan2}, which discriminates between real and fake images generated with proxy masks. The overall training term for the semantic mapper comprises adversarial loss, reconstruction loss, and R1 regularization.

\vspace{-5mm}
\begin{align} 
    & \mathcal{L}_{\text{total}} = \lambda_{\text{adv}}\mathcal{L}_{\text{adv}} + \lambda_{\text{rec}}\mathcal{L}_{\text{rec}} + \lambda_{\text{R1}}\mathcal{L}_{\text{R1}} 
    \label{eqn:semantic_mapper}
\end{align}

Through this Semantic Guide, we have enabled our generator to capture and integrate the desired semantic information effectively. Consequently, the synthesized images produced by our method closely align with the user's intended semantic content.

\section{Experiments}

% 순서: sis를 먼저 보여주는 방향
% full supervision을 사용했을 때 성능이 떨어지는 것을 limitation언급 

% 

We have conducted a range of experiments to highlight the adaptability of our approach across diverse datasets, including CelebAMask-HQ \cite{celebahq}, LSUN Church, LSUN Bedroom \cite{lsun}, FFHQ \cite{stylegan}, and AFHQ \cite{stargan2}. Moreover, we used these datasets for conditional image synthesis under various conditions, including segmentation masks, sketches, and user scribbles. Regarding model architecture, we employed the generator and discriminator of StyleGAN2 \cite{stylegan2} as a pretrained model. Furthermore, we used an architecture similar to SPADE \cite{spade} and a simple UNet architecture as a mapper architecture for segmentation masks and different conditions, respectively. For more detailed information about these experiments and settings, please refer to the Appendix. In addition, we include an ablation study in the Appendix that shows the validity of our proposed loss function and the appropriate number of clusterings.
%% 평가 어떻게 했는지는 appendix로 빼거나 아니면 각각 experiment에서 설명해주는 편이 좋을듯함. 실험마다 데이터도 달라서 여기 적으려니 장황해지는듯.
%% 아래 이상

\subsection{Semantic image synthesis}

\subsubsection{Comparison to few-shot semantic image synthesis method}

We present a comparison with LinearGAN \cite{lineargan}, a method specifically tailored for few-shot semantic image synthesis utilizing a pretrained generator. LinearGAN optimizes the latent vector, which most closely matches the shape of the mask, using few-shot semantic segmentation for semantic image synthesis. Our method surpasses LinearGAN in semantic image synthesis performance with only a single imput mask. To provide an accurate comparison with LinearGAN, we also conducted comparisons in a few-shot setting. 

\begin{figure}
  \centering
  \includegraphics[width=\linewidth]{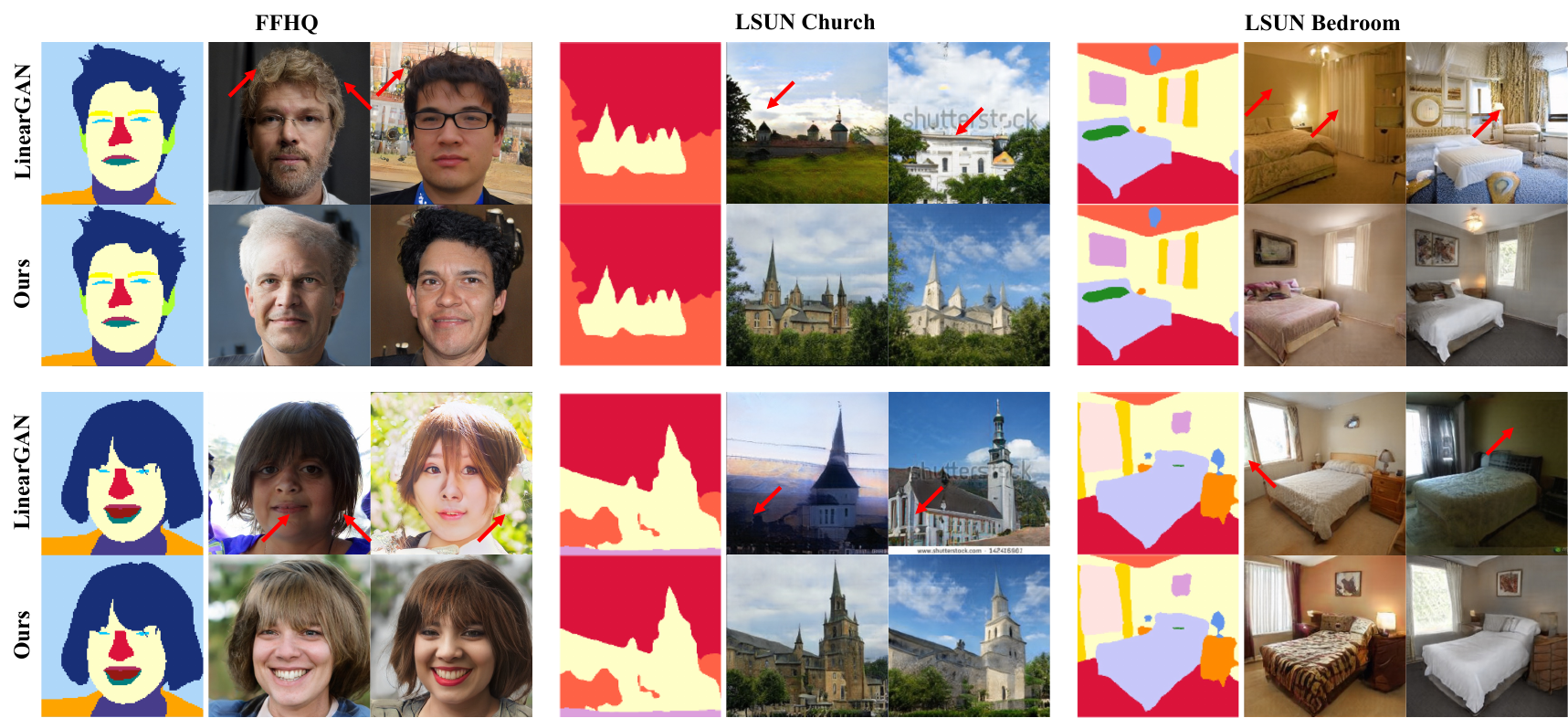}
  \caption{Comparison with LinearGAN when receiving the same mask. LinearGAN is an indirectly optimized method that adjusts the mask through optimization, while our approach is designed to create images that better fit the given mask. Our method performs better by producing more accurate images that align with the masks. Both LinearGAN and Ours use a single mask to train each model.}
  \label{fig:few_shot}
  \vspace{-3mm}
\end{figure}

\fref{fig:few_shot} illustrates the difficulty of LinearGAN in accurately reflecting the overall mask structure. While LinearGAN can generate images that align with the input mask for datasets with less shape diversity, such as human faces, it struggles with LSUN church or LSUN bedroom because of the diverse shape of the datasets. This challenge arises as LinearGAN often finds it difficult to identify a latent closely resembling the shape. Consequently, it fails to reflect intricate shapes effectively. In contrast, our method adjusts the feature maps to correspond with the shape of the input mask, demonstrating better results with detail.

The quantitative comparison, as seen in \tref{tab:lineargan_miou}, utilizes the mIoU between the input mask and the mask generated by a pretrained segmentation network. This metric evaluates the accuracy of the model in reflecting the mask. Our method, surpassing LinearGAN across all datasets, shows its precision in reflecting the input mask, even with limited data. Remarkably, this superiority is maintained irrespective of the number of input masks. This advantage is both qualitatively and quantitatively evident, validating the effectiveness of our approach in few-shot semantic image synthesis.

%Please add the following packages if necessary:
%\usepackage{booktabs, multirow} % for borders and merged ranges
%\usepackage{soul}% for underlines
%\usepackage[table]{xcolor} % for cell colors
%\usepackage{changepage,threeparttable} % for wide tables
%If the table is too wide, replace \begin{table}[!htp]...\end{table} with
%\begin{adjustwidth}{-2.5 cm}{-2.5 cm}\centering\begin{threeparttable}[!htb]...\end{threeparttable}\end{adjustwidth}
\begin{table}[t]\centering
% \scriptsize
\small
{\small
\begin{tabular}{lrrrrr}\toprule
\textbf{N} &\textbf{} &\multicolumn{3}{c}{mIoU} \\\cmidrule{3-5}
&\textbf{Method} &Church &Bedroom &FFHQ \\\midrule
1 &LinearGAN &16 ± 1.4 &17.5 ± 2.0 &37.2 ± 0.8 \\
&Ours &\textbf{23.8 ± 4.4} &\textbf{35.6 ± 4.0} &\textbf{49.8 ± 1.0} \\\midrule
4 &LinearGAN &18 ± 1.3 &21.6 ± 0.9 &39.1 ± 0.5 \\
&Ours &\textbf{27.6 ± 2.09} &\textbf{44.3 ± 2.4} &\textbf{54.3 ± 1.18} \\\midrule
8 &LinearGAN &19.6 ± 0.5 &21.7 ± 0.8 &39.4 ± 0.9 \\
&Ours &\textbf{28.7 ± 1.23} &\textbf{44.7 ± 3.0} &\textbf{54.4 ± 1.7} \\\midrule
16 &LinearGAN &20.4 ± 0.6 &22.3 ± 0.4 &40 ± 0.2 \\
&Ours &\textbf{28.8 ± 1.08} &\textbf{45.7 ± 0.9} &\textbf{56.7 ± 1.19} \\
\bottomrule
\end{tabular}
}
\vspace{2mm}
\caption{mIoU results of LinearGAN and Ours on various datasets. N means number of mask images used in few-shot setting.}\label{tab:lineargan_miou}
\vspace{-8mm}
\end{table}

\subsubsection{Comparison to full-shot method}

Furthermore, we carried out both quantitative and qualitative assessments of our proposed method and the baseline approaches with CelebAMask-HQ \cite{celebahq}, LSUN Church, LSUN Bedroom \cite{lsun}, and FFHQ \cite{stylegan}.

\begin{figure}
  \centering
  \includegraphics[width=0.95\linewidth]{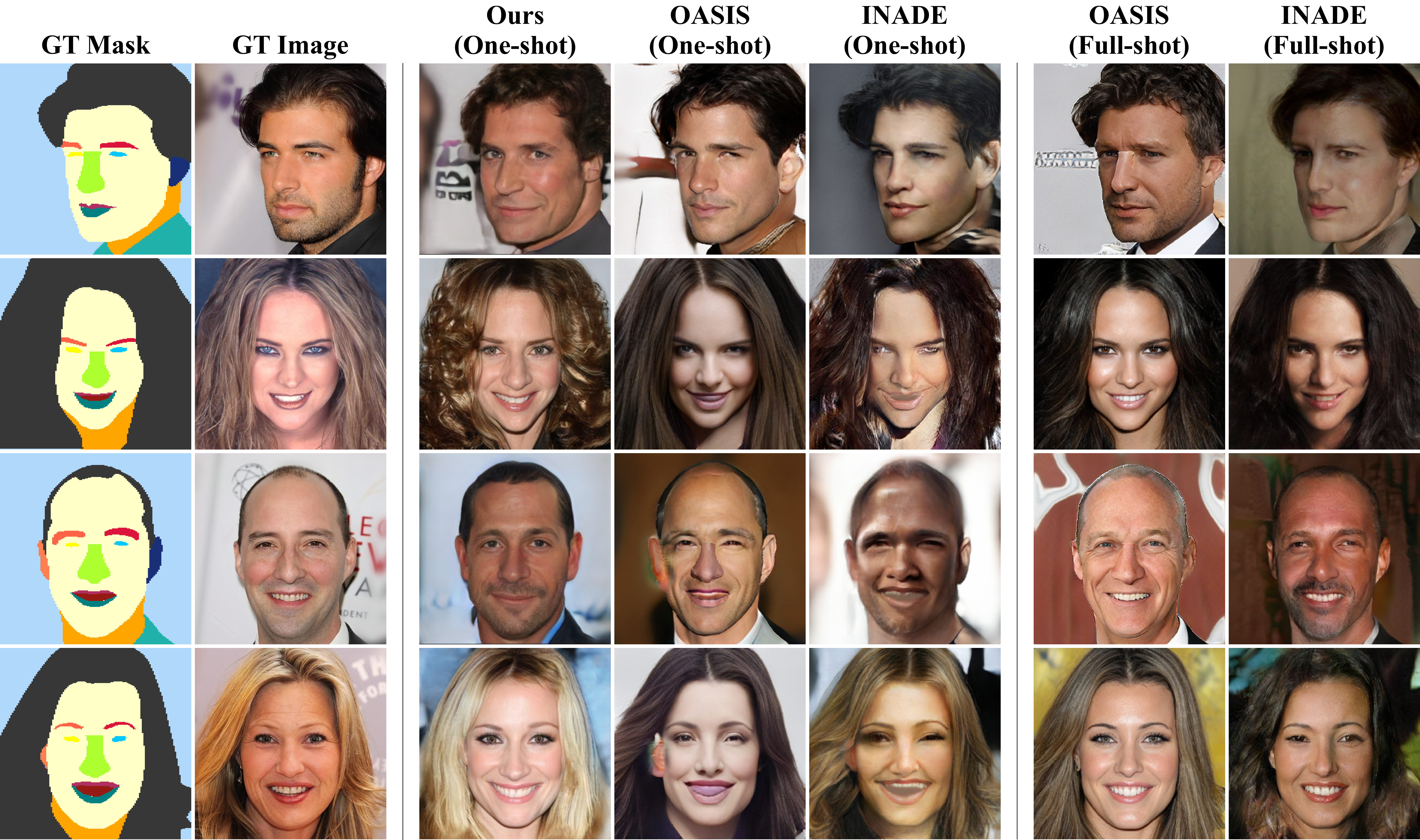}
  \caption{The comparison results between our method and supervised models using the same mask (Including one-shot setting). Compared to existing SIS methods, Ours produces results that are closer to the ground truth. Additionally, Our result shows the most natural-looking output.}
  \label{fig:vs_supervision}
  \vspace{-3mm}
\end{figure}

% one shot vs full shot
As illustrated in \fref{fig:vs_supervision}, we adopted the RepurposeGAN \cite{repurposegan} approach to generate one-shot data pairs for comparison under a one-shot setting. While the results from competitors frequently display unnatural artifacts, our method yields more natural and closer-to-ground truth outcomes. Also, as depicted in \tref{tab:quantitative}, our method has a higher mIoU than existing techniques with one-shot training data pairs. The reason is that our approach excels in accurately segmenting large mask regions. Our method shows lower mIoU than the conventional models trained with full-shot 28K image-annotation pairs because IoU decreases for fine-grained classes that are not included in the one-shot setting, such as accessories with low frequencies. However, when it comes to the FID, our work significantly outperforms the one-shot settings. It is typically superior to the full-shot methods, even with just a single annotation. In addition, we have included results for other metrics, such as LPIPS and Precision and Recall, in the Appendix.

%Please add the following packages if necessary:
%\usepackage{booktabs, multirow} % for borders and merged ranges
%\usepackage{soul}% for underlines
%\usepackage[table]{xcolor} % for cell colors
%\usepackage{changepage,threeparttable} % for wide tables
%If the table is too wide, replace \begin{table}[!htp]...\end{table} with
%\begin{adjustwidth}{-2.5 cm}{-2.5 cm}\centering\begin{threeparttable}[!htb]...\end{threeparttable}\end{adjustwidth}

\begin{table}[!htp]\centering
%\scriptsize
\small
{\small
\begin{tabular}{lrrrrrrr}\toprule
&FID &mIoU(vs gt mask) &Acc(vs gt mask) \\\midrule
% Pix2PixHD &38.5 & &76.1 &x &93.06 &x & \\
% SPADE &29.2 & &75.2 &x &93.26 &x & \\
SEAN  &28.1 & 75.9 &92.3 \\
OASIS  &20.2 & 74.0 &91.5  \\
INADE  &21.5 & 74.1 &93.2  \\
SDM\cite{sdm}  &18.8 & 77.0 &n/a \\\midrule
OASIS (1-shot) &25.5 &45.5 &82.7 \\
INADE (1-shot) &28.1 & 44.3 &83.8  \\
Ours (1-shot) &18.5 &53.1 &88.2  \\
\bottomrule
\end{tabular}
}
\vspace{3mm}
\caption{Quantitative comparison with SEAN, OASIS(1-shot / full-shot), INADE(1-shot / full-shot), SDM. mIOU and Acc are measured by comparing with the ground truth mask.}\label{tab:quantitative}
\vspace{-5mm}
\end{table}

% \cjw{우리꺼 Full shot했을때 FID 15.37, mIoU 60.92로 올라간다라는 내용을 추가 - 표에 넣기엔 다른 full-shot보다는 떨어져서 애매함}

% For instance, low-frequency classes like earrings or hats may not be accurately segmented by our method, leading to incorrect labeling of regions, such as the creation of hair masks instead of these accessories. 

\begin{figure}
  \centering
  \includegraphics[width=0.95\linewidth]{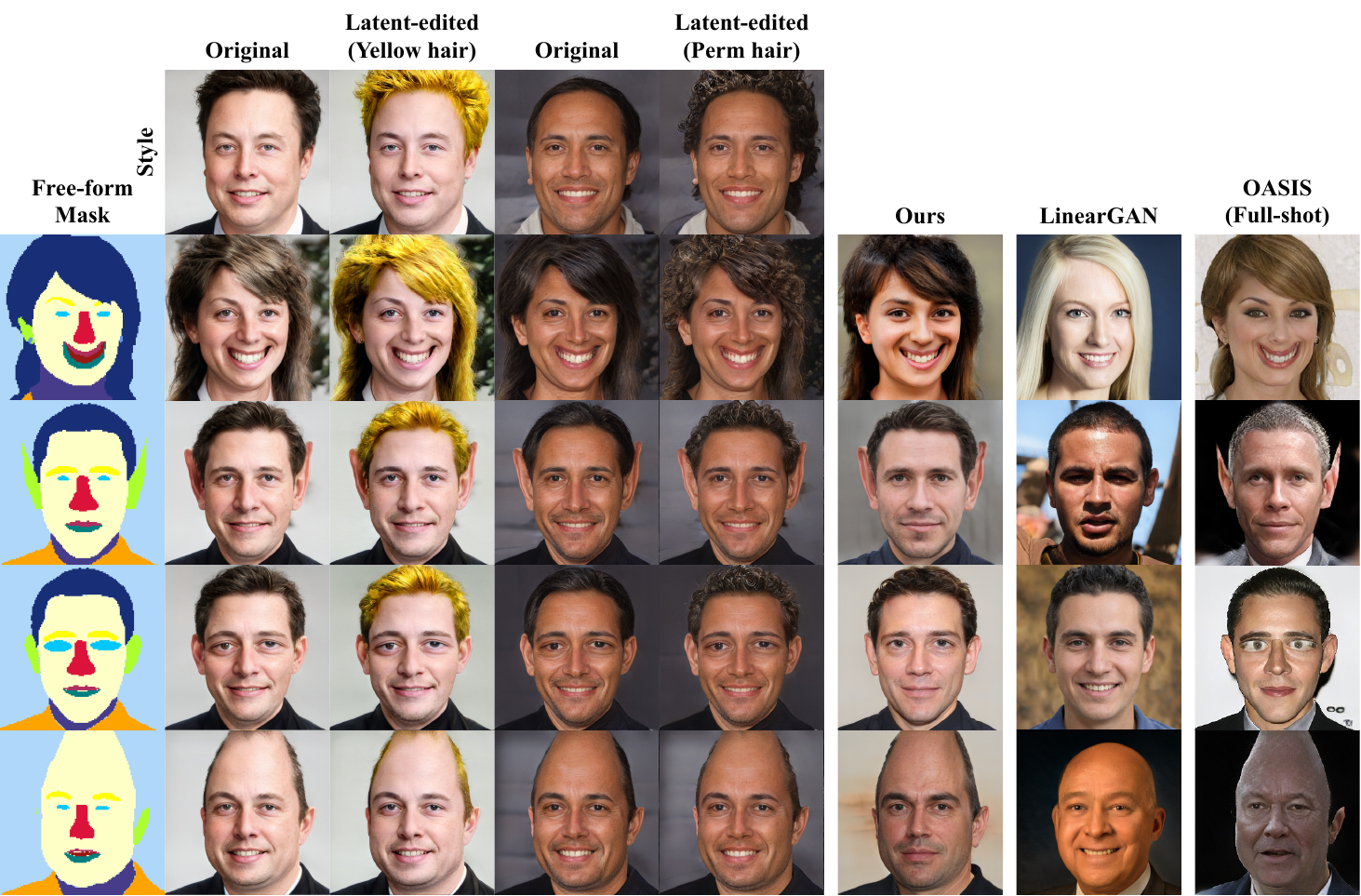}
  \caption{Result of image synthesis with free-form mask. Images in the first row are inverted images in StyleGAN and their edited images with the HairCLIP \cite{hairclip}. Images are converted to the spatial shapes of both masks and images in the first column while preserving styles and identities. We also show a comparison with LinearGAN \cite{lineargan} and OASIS \cite{oasis} in the last three columns. For comparison we sample the images with random noise for all models}
  \label{fig:free_form}
  \vspace{-3mm}
\end{figure} 

\subsection{Free-form image manipulation}

In this subsection, we show the flexibility of our proposed method in facilitating free-form image manipulation. Our approach underscores its capability of handling masks that deviate substantially from typical training distributions, such as oversized human ears and big oval eyes. \fref{fig:free_form} illustrates the image synthesis results when such unconventional masks are used as inputs. Even in cases where the masks deviated from the typical human form, our method successfully synthesized results that corresponded appropriately to the provided masks. In contrast, when LinearGAN \cite{lineargan} and OASIS~\cite{oasis} receive a mask that falls outside the typical human range as a shape guide, they fail to generate results that align with the shape indicated by the mask.

Moreover, the versatility of our method is further emphasized by its compatibility with latent editing techniques such as HairCLIP \cite{hairclip} as shown in \fref{fig:free_form}. While latent editing changes the hairstyle to a yellow color or perm style, our method preserves the shape of images corresponding to given masks. See the Appendix G for additional samples of generated images conditioned with free-form masks.

\begin{figure}[h]
    \centering
    \includegraphics[width=0.95\linewidth]{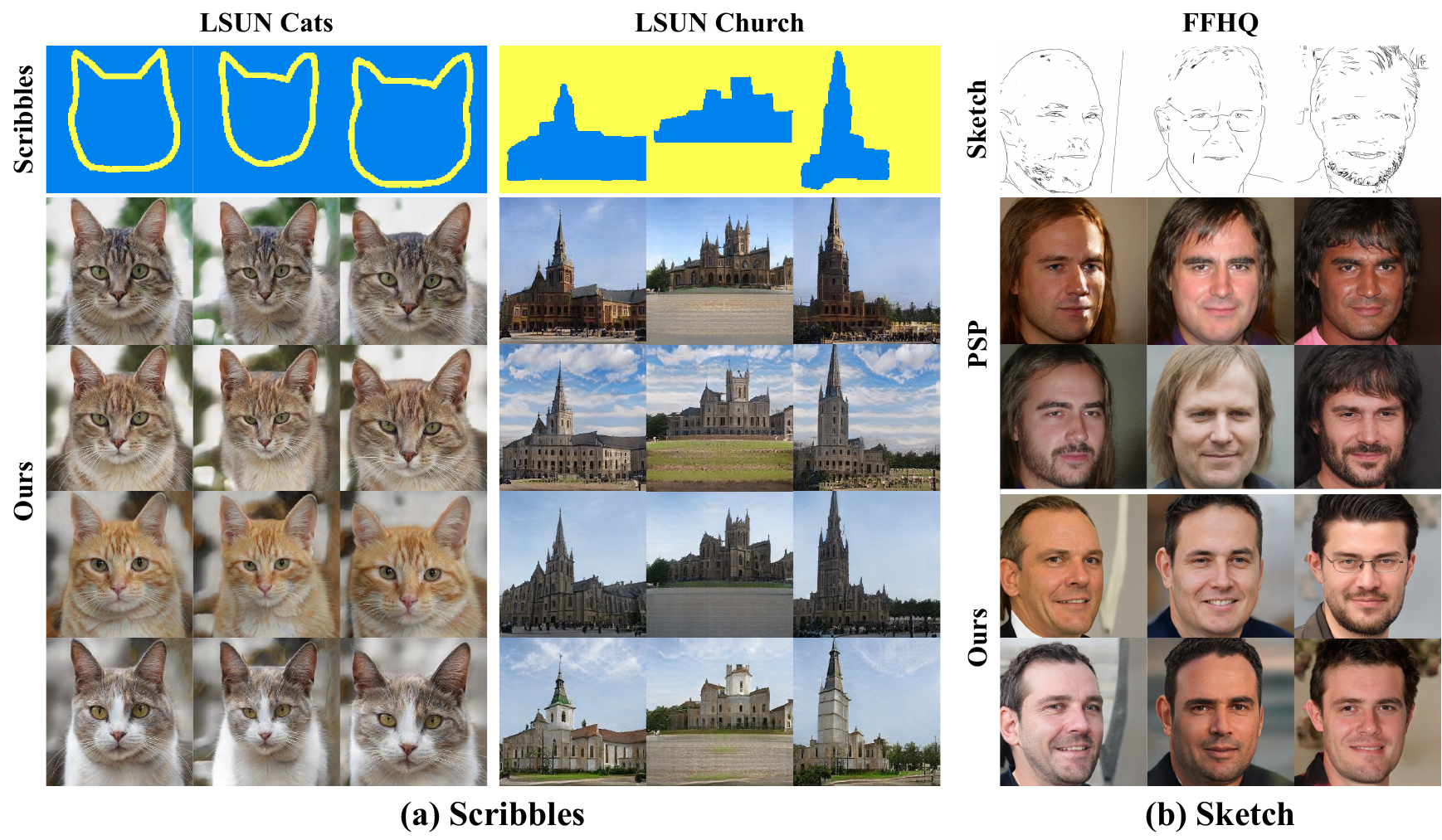}
    \caption{Image synthesis results when diverse types of input masks are entered, such as scribbles and sketches.}
    \label{fig:sketches_scribbles}
    \vspace{-3mm}
\end{figure}

\subsection{Conditional image synthesis using a variety of conditions.}
This part broadens the scope of semantic image synthesis to include just segmentation masks and diverse conditions like sketches and user scribbles. \fref{fig:sketches_scribbles}(a) demonstrates the outcome when simple scribbles are used as an input mask. Remarkably, even with a scribble mask created with just a single line, our method can generate image synthesis that is well-suited to the indicated shape. Moreover, this user-friendly aspect further enhances the usability of our approach. Additionally, we visualize our results and compare them qualitatively with pSp \cite{psp}, a GAN inversion approach demonstrating its ability to invert various inputs into latent codes, not solely real images. As illustrated in \fref{fig:sketches_scribbles}(b), our comparison with pSp, especially in semantic image synthesis using sketch inputs, reveals a significant difference. pSp can generate coarse structures effectively with pSp encoder and style mixing, but it struggles to capture and reproduce the fine details of the input conditions. On the other hand, our method surpasses pSp with more corresponding results with fewer restrictions of matching latent code because of the flexible spatial control of our model using feature maps rearranging. In the Appendix E, we further demonstrate the proficiency of our method in generating images based on various input conditions with implementation details. This additional evidence underscores the versatility of our method and superior performance in conditional image synthesis.

% We compare our method with supervised semantic image synthesis techniques. To create a fair comparison under a 1-shot setting, we utilize the RepurposeGAN \cite{repurposegan} to generate 1-shot data pairs and train competing methods \cite{oasis, inade} using these pairs. A qualitative comparison of the results is presented in Figure \fref{fig:free_form}. While competitor methods often produce results tainted by unnatural artifacts, our approach generates more natural and ground truth-like images. 

% However, as evidenced in Table \tref{tab: quantitative}, our method registers a marginally lower mIOU compared to existing fully-annotated supervised techniques. This disparity primarily arises from the fact that our method excels in accurately segmenting larger, coarser mask regions, but struggles when it comes to smaller, fine-grained classes - especially those that occur infrequently and are not included in the 1-shot setting. This shortfall contributes to significant reductions in quantitative metrics. For instance, low-frequency classes such as earrings or hats may not be accurately segmented by our method, leading to region mislabeling. An example of this mislabeling could be the generation of hair masks in place of these accessories.

% Despite these challenges, our method continues to deliver more realistic images compared to its competitors. This can be attributed to our method's effective utilization of the rich image generation capabilities of the pretrained unconditional generator. 

\begin{figure}
  \centering
  \includegraphics[width=\linewidth]{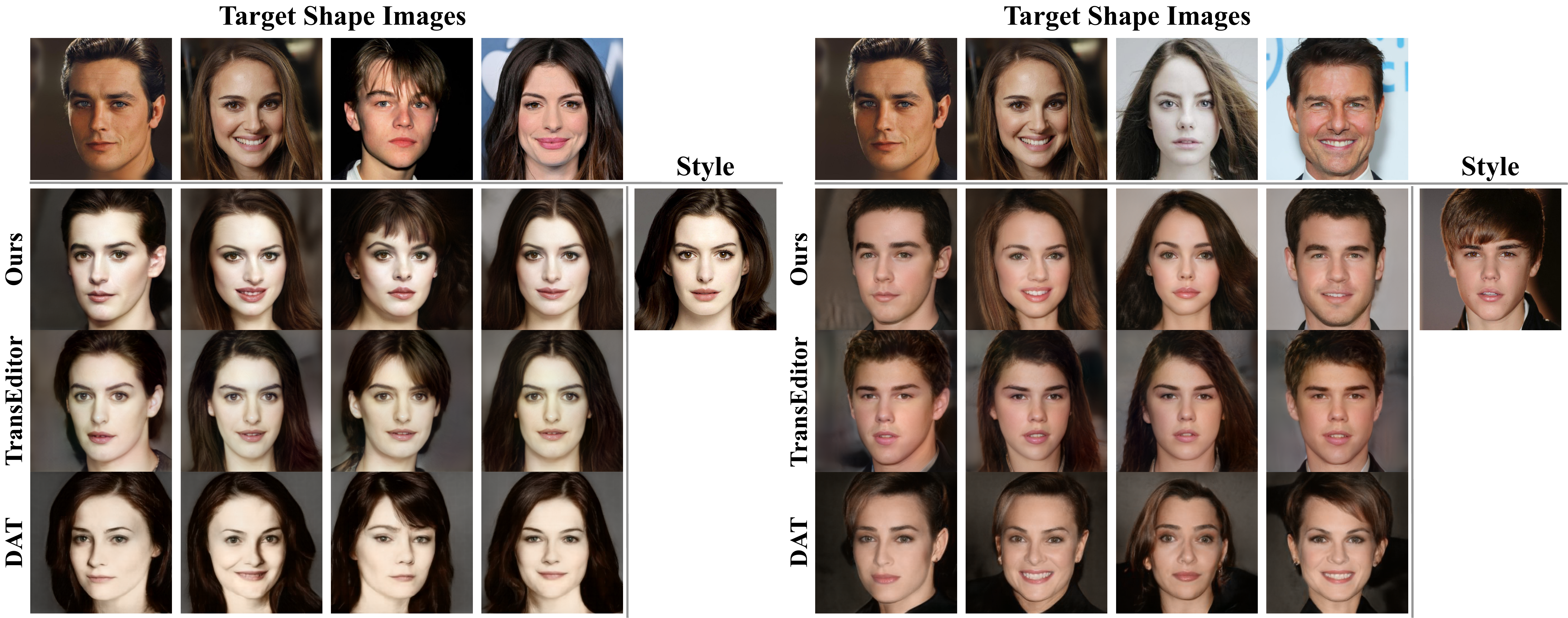}
  \caption{Image generation results with a real image as a guide. The top row consists of images that serve as the style input, while the left column contains target image that determine the desired shape.}
  \label{fig:exampler}
  \vspace{-3mm}
\end{figure} 
%% target을 여러개.

\subsection{Exemplar-guided image generation} 

Beyond the specified conditions, our model produces an output image that has a similar structure to a provided sample. The results of image generation are displayed in \fref{fig:exampler}. The latent codes are derived from real images by utilizing the e4e inversion technique \cite{e4e} that was trained with the CelebAMask-HQ training set. This inverted latent code enables us to obtain a proxy mask of the sample, thereby aligning the generated images with the real image structure.

In \fref{fig:exampler}, it is evident that our generated results better preserve shape in comparison to other exemplar-guided image generations. The results of Transeditor \cite{transeditor} and DAT \cite{diagonalgan} do not preserve the original shape as effectively, with DAT even showing a disparity in style. Although competitors use dual latent spaces to portray exemplar structure more accurately, the structural latent code does not always reflect the shape correctly. In contrast, our method more accurately mirrors the exemplar shape because the Rearranger directly forms the feature maps with the proxy mask.

%우리 방법이 왜 더 좋은지 정도 써줘야할듯 결과해석까지만 간단히 써있음.

%As mentioned in (\secref{sec:related_work}), 

%여기는 reference로 삼을 shape을 real image 혹은 fake image로 사용할 수 있다는 것을 보여주려고 함. (Fully unsupervised) 그래서 어떤 reference 이미지를 받았을때 그에 맞는 shape으로 다른 이미지들을 변경시키는걸 figure로 넣을 것임. + TransEditor, CoordGAN, DAT 정도가 들어갈 수 있음
%%% One shot dependency도 큼.

% \begin{figure}[h]
%   \centering
%   \includegraphics[width=0.85\linewidth]{images/limitation.pdf}
%   \caption{\textbf{Failure cases} of our method. Boxed areas show where the generated images do not match the corresponding masks.} 
%   \label{fig:limitation}
% \end{figure}

\section{Conclusion}

Semantic image synthesis has received significant attention for its ability to reflect intentions on images at a very detailed level, but achieving this requires substantial data. Therefore, we proposed our method as a solution to address this issue. Our method rearranges the features of a pretrained generator to enable detailed control over images, and all elements of our approach are self-supervised, except for a single annotation mapped to human perception. With just one annotation, we achieve precise image control, as demonstrated through experiments in semantic image synthesis with various input conditions. Our method also showed data efficiency in situations where the pixel-level annotation changes. Overall, our approach demonstrated a high level of detail control without the dependence on annotations.
\paragraph{Limitations} Our one-shot results outperform other competitors, but a slight performance gap remains when compared to full-shot outcomes. Therefore, future work will make it possible to outperform to beat full-shot performance. Also, our approach operates at the feature map level, which, while enabling the natural incorporation of masks, needs to achieve complete pixel-level control. For more details, please refer to the Appendix D.

\section*{Acknowledgement}
This work was supported by the National Research Foundation of Korea(NRF) grant (No. 2022R1F1A107624111) funded by the Korea government (MSIT).

% \fref{fig:limitation} shows failure cases of our method because of the feature maps resolution. This provides an opportunity for future work focused on enhancing pixel-level manipulability

% 트랙킹 부분 자연스럽게 고치기 
% ur method is versatile across various applications such as free-form spatial editing of real images, sketch-to-photo, and even scribble-to-photo.

% \paragraph{Limitation} Our one-shot results outperform those of other competitors, but a slight performance gap remains when compared to full-shot outcomes. Therefore, future work will make it possible to outperform to beat full-shot performance. And also, our approach operates at the feature map level, which, while enabling the natural incorporation of masks, falls short of achieving complete pixel-level control. \fref{fig:limitation} shows failure cases of our method because of the feature maps resolution. This provides an opportunity for future work focused on enhancing pixel-level manipulability. 
%Finally, a noteworthy consideration is the computational demand of our method. Despite its relatively low parameter count, which allows for all training to be completed within a day even on a personal GPU RTX 3090, the attention operation conducted on high-resolution feature maps incurs substantial computation. Therefore, future work could also focus on improving computational efficiency, possibly inspired by methods such as Vision Transformers (ViT)\cite{vit}, which operate attention at the patch level, promising an exciting direction for future advancements.
{\small
    \bibliography{egbib}
    \bibliographystyle{plainnat}
}

\clearpage

% \renewcommand{\thesection}{\Alph{section}}

% \section{Appendix}

\begin{appendix}

% \maketitle

\renewcommand{\thetable}{S\arabic{table}}
\renewcommand{\thefigure}{S\arabic{figure}}
\setcounter{figure}{0}
\setcounter{table}{0}

\section{Implementation details}
\subsection{Architectural designs}
\begin{figure}[h]
  \centering
  \includegraphics[width=0.95\textwidth]{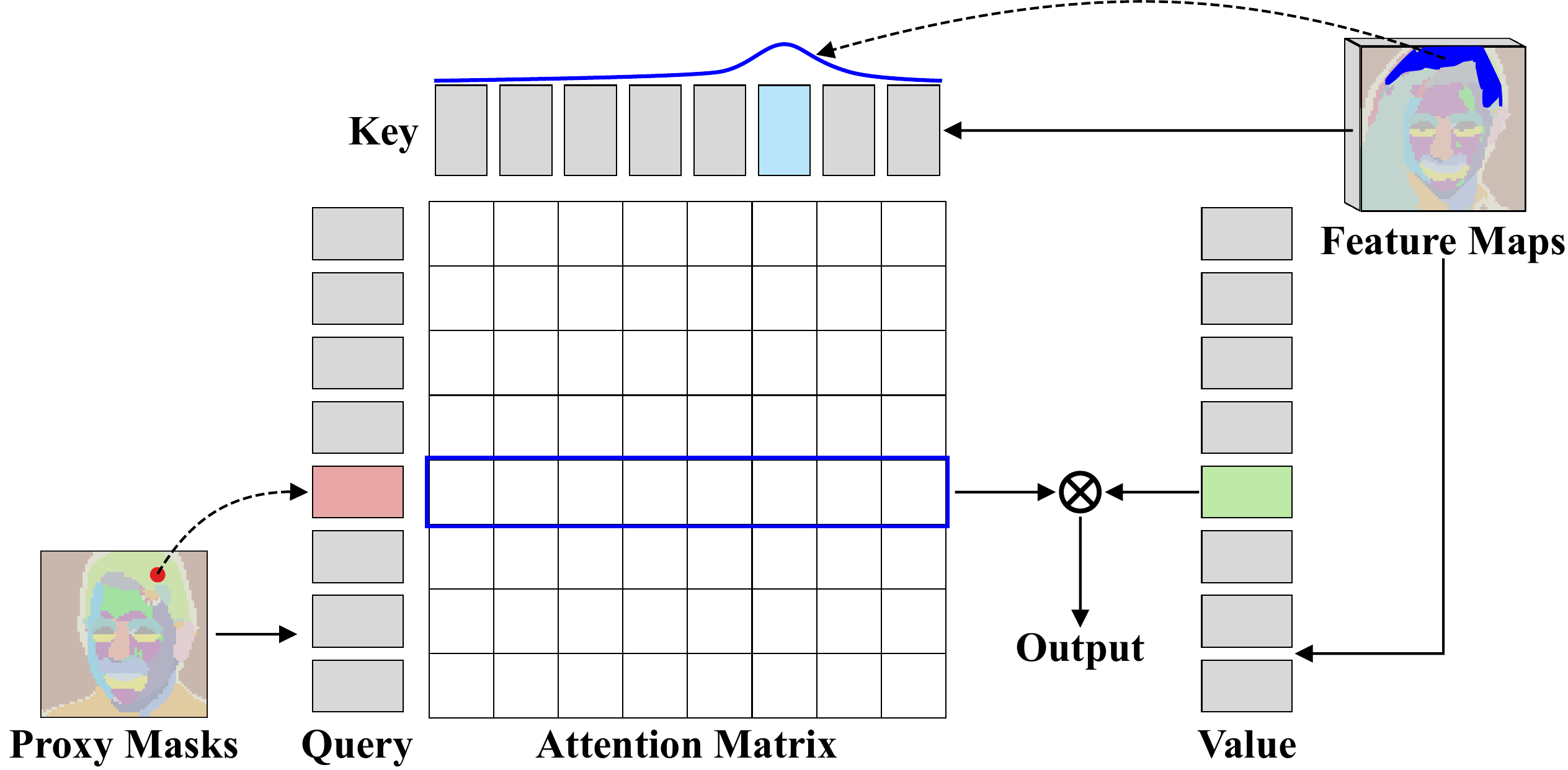}
  \caption{\textbf{Illustration of the cross-attention operation.} The module computes an attention matrix using the query (proxy mask) and key (feature maps), enabling the rearrangement of the value (feature maps) according to the shape of the query. }
  \label{fig:attention}
\end{figure}

We provide additional details of the architectural designs for our rearranging network and semantic mapper. For the rearranging network, we incorporate a single head cross-attention module, which is surrounded by two 4-layer residual blocks on either side. In \fref{fig:attention}, the cross-attention module operates by computing an attention matrix from the query (the proxy mask) and the key (feature maps). This process enables the value (feature maps) to be rearranged (through a weighted sum) to align with the form of the query, thereby reflecting their strong correspondence. To reflect dissimilarity between different pixels in a mask, we add a sinusoidal position encoding before the residual block that precedes the module.
% By doing so, each pixel of the mask received class and position encoding, thus enhancing the correspondence with the matching pixel in the feature maps.

For the semantic mapper, we adopt the architecture of OASIS \cite{oasis}, which takes the input condition and generates a $64^2$ resolution output proxy mask, except input noise. This method ensures a high level of incorporation of the input mask in the proxy mask. The input noise is removed because its stochasticity slows down the training.
% We found that the stochasticity of OASIS slowed down convergence, so we adopted a structure that excluded 3D noise. 
However, when the input condition is not a semantic segmentation mask, it is not appropriate to use a SPADE layer \cite{spade}. As a solution, we added a U-Net structured decoder to the encoder which has same structure with discriminator of the StyleGAN2 \cite{stylegan2}, allowing the shape of the input mask to be preserved all the way to the output proxy mask.

\subsection{Training details}

In our training setup, we follow a similar approach to that of StyleGAN2\cite{stylegan2}, where the learning rate is set to 0.002 for both the generator and discriminator. Additionally, the weight for $\text{R}_{1}$ regularization is set to 10.
% was applied with a weight of 10. 
% In consideration of the resolution of our training data, we used a The batch size is set to 4 for $256^2$ datasets, and 2 for $512^2$ and $1024^2$ datasets, respectively.

% We implemented training in a self-supervised manner, 
% We design a self-supervised training, 
We prepare the proxy masks from the centroids obbtained by clustering feature maps of 256 sample images. The size of the feature maps is $64^2$.
% wherein proxy masks are constructed from centroids obtained through clustering of feature maps of 256 sample images, each of a resolution of $64^2$. 
To achieve this, we employ K-means clustering~\cite{edit}. Given the need for balancing between high correspondence and image quality, we empirically set the weights of our loss terms. Specifically, we set $\lambda_{\text{mask}}$ to 1.0 and $\lambda_{\text{self}}$ to 10.0. To prevent shortcuts of self-reconstruction loss, we apply a random horizontal flip augmentation. Furthermore, we alternate training with a proxy mask generated from the same noise as the feature maps used for self-reconstruction and a proxy mask generated from random noise in each iteration.

 % To achieve this, we employed K-means clustering, as referenced in \cite{edit}.

The training for our rearrangement network was conducted using the CelebAMask-HQ \cite{celebahq}, LSUN Church, and LSUN Bedroom datasets \cite{lsun}, each at a resolution of 256. We adopted a strategy where $\mathcal{L}_{\text{self}}$ and $\mathcal{L}_{\text{mask}}$ were computed for every iteration during the initial 100k iterations, whereas $\mathcal{L}_{\text{adv}}$ was computed once every five iterations. Subsequently, all three loss components were computed for each iteration over the next 40k iterations. When training with the 1024 resolution FFHQ \cite{stylegan} and 512 resolution AFHQ\cite{stargan2} datasets, we computed $\mathcal{L}_{\text{self}}$ and $\mathcal{L}_{\text{mask}}$ at each iteration for the first 150k iterations, with $\mathcal{L}_{\text{adv}}$ being evaluated once every 5 iterations. Subsequently, all three loss terms were computed for each iteration over the following 65k iterations.

For the training of the semantic mapper, we initially trained for the first 100k iterations using only $\mathcal{L}_{\text{recon}}$ with $\lambda_{\text{self}}$ to 10.0. Then, for the subsequent 10k iterations, $\mathcal{L}_{\text{adv}}$ was added to calibrate the generated image within its domain, during which we adjusted $\lambda_{\text{recon}}$ to 100.0. It should be noted that all $\lambda_{\text{adv}}$ values used in the mapper and rearranging network were consistently set to 1.0.

% As for the training time, we completed the entire process
The entire training completes within a day for data at a resolution of 256 using an NVIDIA RTX 3090 GPU. For higher resolution configuration ($1024^2$), the training completes within two days. This efficient training process highlights the feasibility of our approach for practical applications.

\section{Ablation study}

\subsection{Proposed loss function}

\begin{figure}[h]
  \centering
  \includegraphics[width=\textwidth]{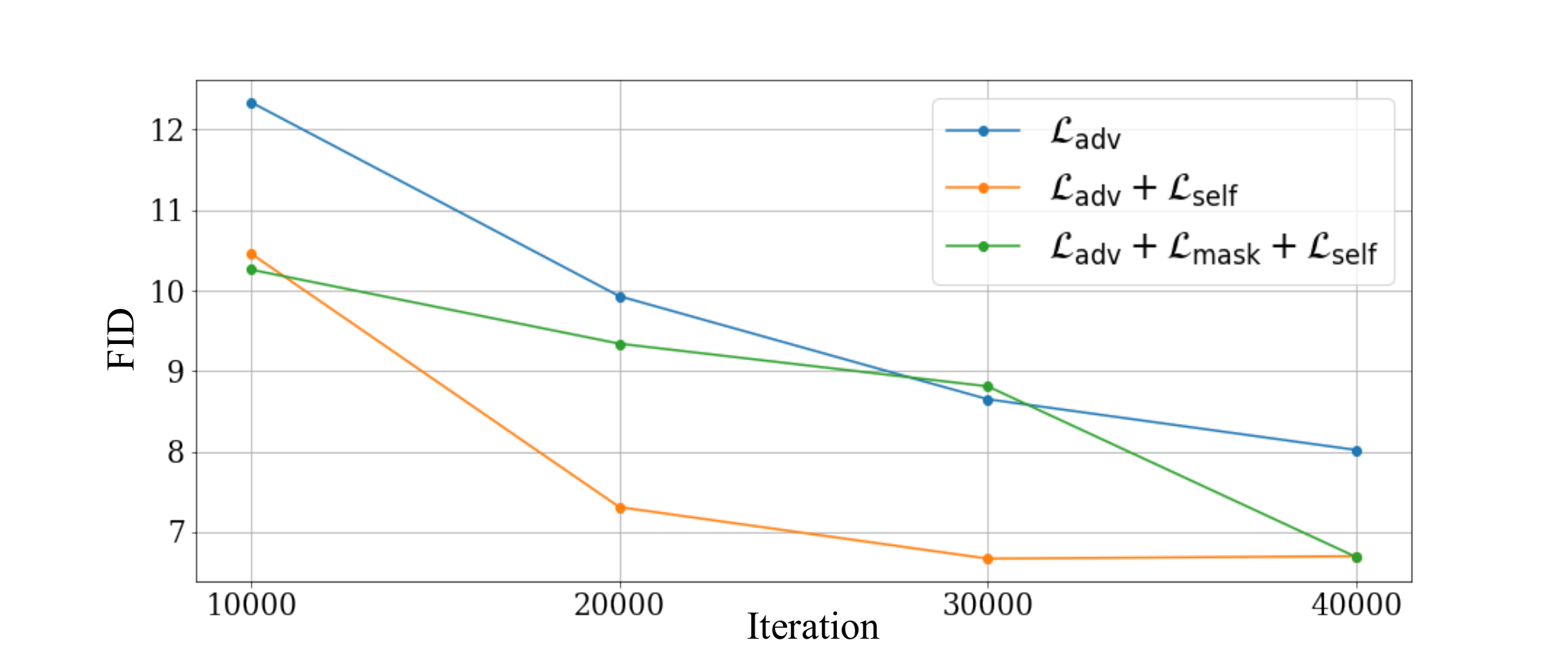}
  \caption{\textbf{Quantitative studies to demonstrate the impact of additional losses introduced in our method} $\mathcal{L}_{\text{self}}$ leads to a rapid decrease in FID, thus demonstrating its contribution to achieving fast convergence.} 
  \label{fig:ablation_fid}
\end{figure}

%Adversarial loss만 써도 우리의 모듈이 work하는 것을 보임 하지만 Correspondece 및 Style 보존 능력이 매우 떨어짐. 
%Self-Reconstruct loss(본문에서 terminology 주의 할 것) 추가하여 위의 결점을 보완하며 또한 더 빠른 수렴을 보여 줌 (아직 확실하지 않음.fid figure뽑아봐야함).
% 다만, Self 만 사용한 경우 본문과 같이 proxy mask를 무시하는 경향이 생기는데 이것을 방지하기 위한 mask loss를 도입하여 아래와 같은 결과를 확인할 수 있음.실험은 celeba에서 이루어짐.

\begin{figure}[t]
  \centering
  \includegraphics[width=\textwidth]{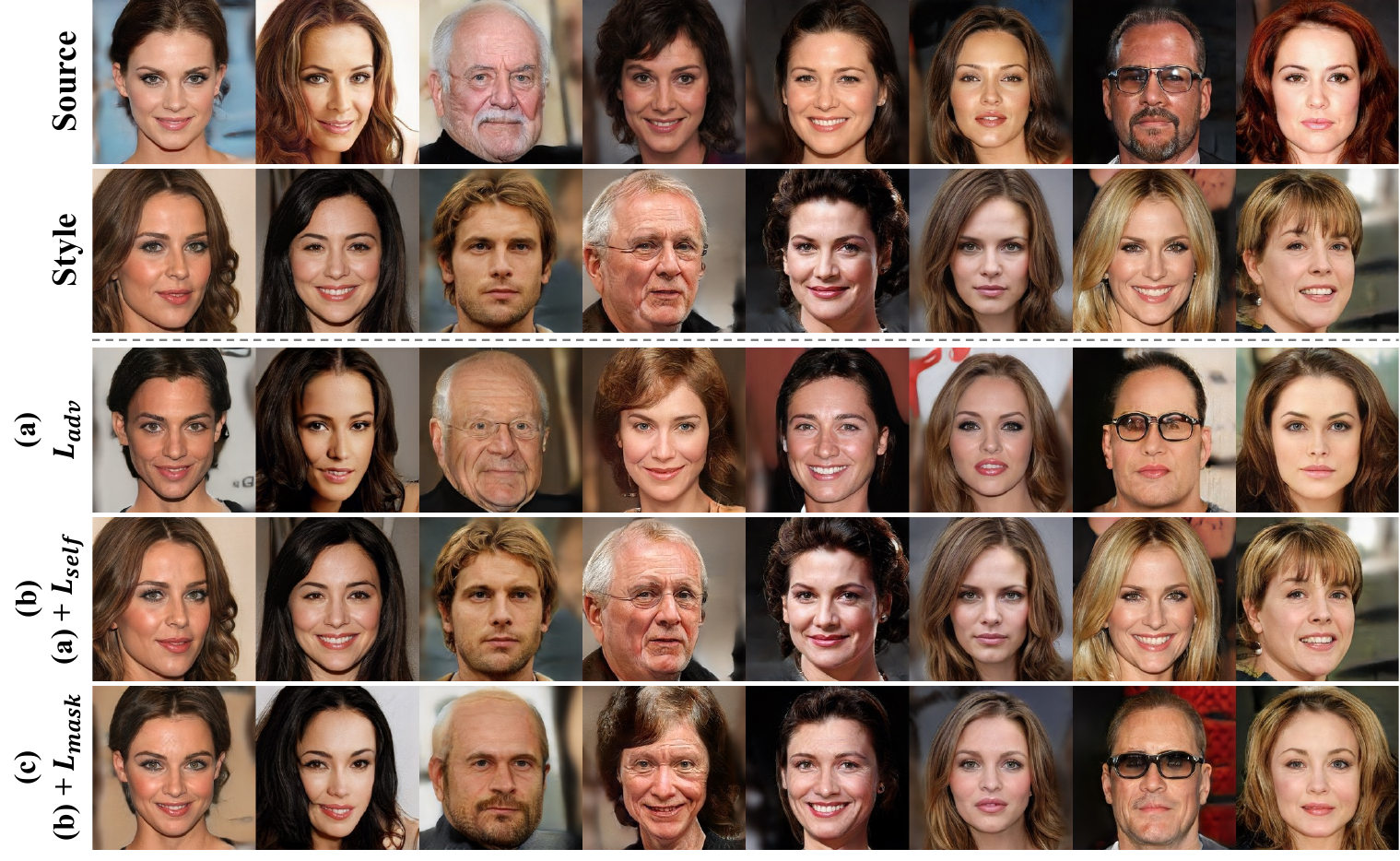}
  \caption{\textbf{Qualitative studies to demonstrate the impact of additional losses introduced in our method} (a) $\mathcal{L}_{\text{loss}}$ contributes to producing realistic results. (b)$\mathcal{L}_{\text{self}}$ increases the preservation rate of style attributes but tends to ignore the shape of source image. (c)$\mathcal{L}_{\text{mask}}$ effectively reflects the shape of the source image.} 
  \label{fig:ablation_ffhq}
\end{figure}

To demonstrate the influence of the additional losses introduced in our method, we provide both quantitative and qualitative ablations in \fref{fig:ablation_fid} and \ref{fig:ablation_ffhq}, respectively. We conducted these experiments on the CelebA dataset, comparing three scenarios: using only $\mathcal{L}_{\text{adv}}$, adding $\mathcal{L}_{\text{self}}$, and using all three losses, including $\mathcal{L}_{\text{mask}}$. The Fréchet Inception Distance (FID) is measured at each iteration, and the degree of attribute preservation is demonstrated through a curation of the images produced at the final convergence point. When evaluating the FID , we generated a dataset consisting of 28,000 images by employing two latent codes. One latent code was used for generating proxy masks, while the other code was utilized for defining the desired style. These generated images were then compared with a dataset of 28,000 training images for FID calculation.

As evident from the plot in Figure \ref{fig:ablation_fid}, the presence or absence of $\mathcal{L}_{\text{self}}$ significantly affects the speed of convergence. This is because the self mask serves as a guide for the rearranging network, particularly in the early stages of training. However, if only $\mathcal{L}_{\text{self}}$ is incorporated, the network tends to disregard the proxy mask, favoring the restoration of its original state, as illustrated in the corresponding figure.

Thus, the inclusion of $\mathcal{L}_{\text{mask}}$ serves to counterbalance this trend, ensuring that the proxy mask is not ignored and is effectively utilized in the image generation process. These results demonstrate the critical role and effectiveness of our proposed loss components in achieving high-quality, attribute-preserving image generation.

\subsection{Number of clusters}

\begin{figure}[h]
  \centering
  \includegraphics[width=\textwidth]{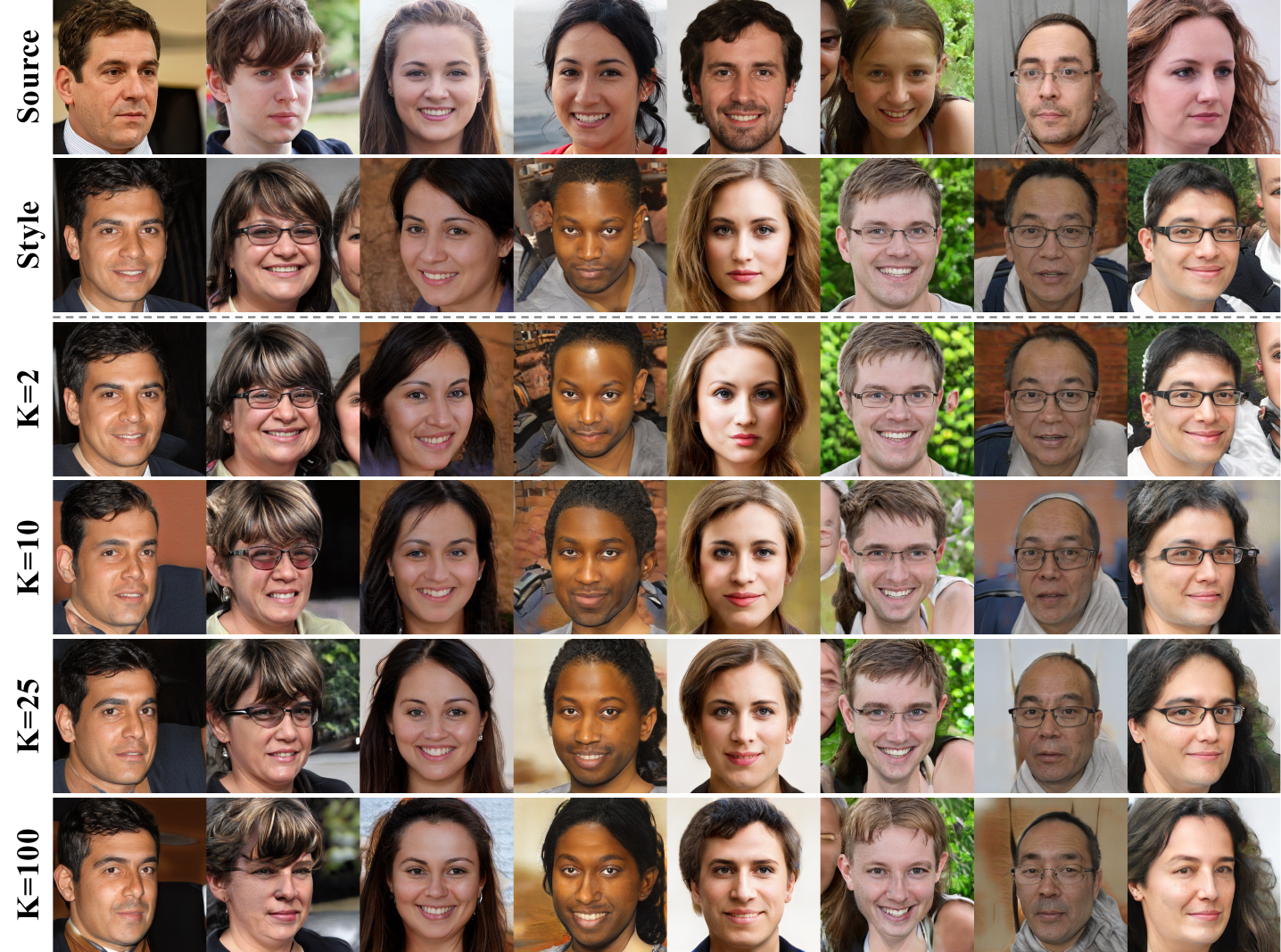}
  \caption{\textbf{Qualitative studies to demonstrate the impact of number of clusters} As the number of clusters increases, more detailed semantic parts, such as glasses, can be separated. K denotes number of clusters.} 
  \label{fig:ablation_ffhq_K}
\end{figure}

The number of clusters significantly impacts the shape of the proxy mask and, consequently, the generated image. Therefore, we carry out a qualitative ablation study on how the number of clusters influences the generated image, the results of which are presented in Figure \ref{fig:ablation_ffhq_K}.

For this evaluation, we generate images by applying rearrangement using the feature map and the proxy mask obtained from random noise. Our results, presented in Figure \ref{fig:ablation_ffhq_K}, show that as the number of clusters increases, detailed elements like glasses are more accurately reflected in the proxy mask. This improvement can be attributed to the increased capacity to capture finer image details as the number of classes in the proxy mask escalates.

However, when the number of clusters is low, we noticed that the influence of the position embedding becomes dominant over the class embedding, resulting in the proxy mask being overlooked and the image not transforming as intended.

Nonetheless, caution is warranted when overly increasing the number of clusters. Beyond a certain point, the proxy mask begins to diverge from the semantics understood by humans, which not only confuses the semantic mapper but also hampers generating a suitable proxy mask from the input mask. Therefore, for the convenience of our experiments, we decided to fix the number of clusters at 25.

\section{Additional quantitative comparison}

\subsection{Additional diversity metric }

Additionally, we assess result diversity using the LPIPS metric. LPIPS diversity of our method surpasses INADE and OASIS, but there is no clear winner when combining precision and recall. \tref{tab:additional_eval}, summarizes these metrics. To measure LPIPS, we follow INADE\cite{inade}, evaluating overall diversity by generating 10 image groups with random noise and calculating diversity scores between two groups. We compute 10 scores, averaging them to reduce fluctuation from random sampling.

\subsection{Training time comparison}

\begin{figure}[h]
  \centering
  \includegraphics[width=\textwidth]{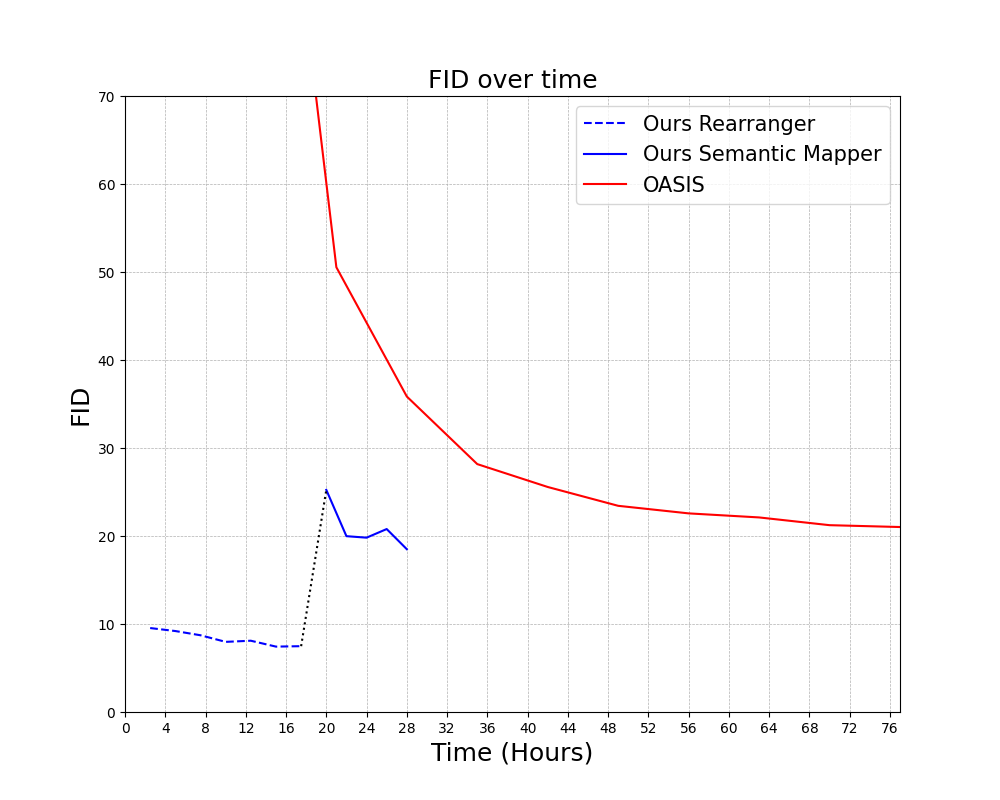}
  \caption{FID of generated images when time passes. The result shows that our work converges faster than OASIS.} 
  \label{fig:fid_score}
\end{figure}

To demonstrate the efficiency of our model, we evaluated our model to OASIS \cite{oasis} using the FID over time. As depicted in Figure \ref{fig:fid_score}, training Rearranger and Semantic Mapper consecutively shows a faster convergence speed than our competitor. Furthermore, if we have trained Rearranger, our model only requires the retraining of the Semantic Mapper when there is a change in the target semantic segmentation mask classes. 

\begin{table}[!htp]\centering
%\scriptsize
\small
{\small
\begin{tabular}{lrrrrrrr}\toprule
&LPIPS &Precision &Recall \\\midrule
Ours  &0.45 & 0.76 &0.22 \\
INADE  &0.36 & 0.86 &0.19  \\
OASIS  &0.30 & 0.69 &0.36  \\
\bottomrule
\end{tabular}
}
\vspace{3mm}
\caption{Additional quantitative comparison with INADE and OASIS.}\label{tab:additional_eval}
\vspace{-5mm}
\end{table}

\section{Failure cases}
\begin{figure}[h]
  \centering
  \includegraphics[width=\textwidth]{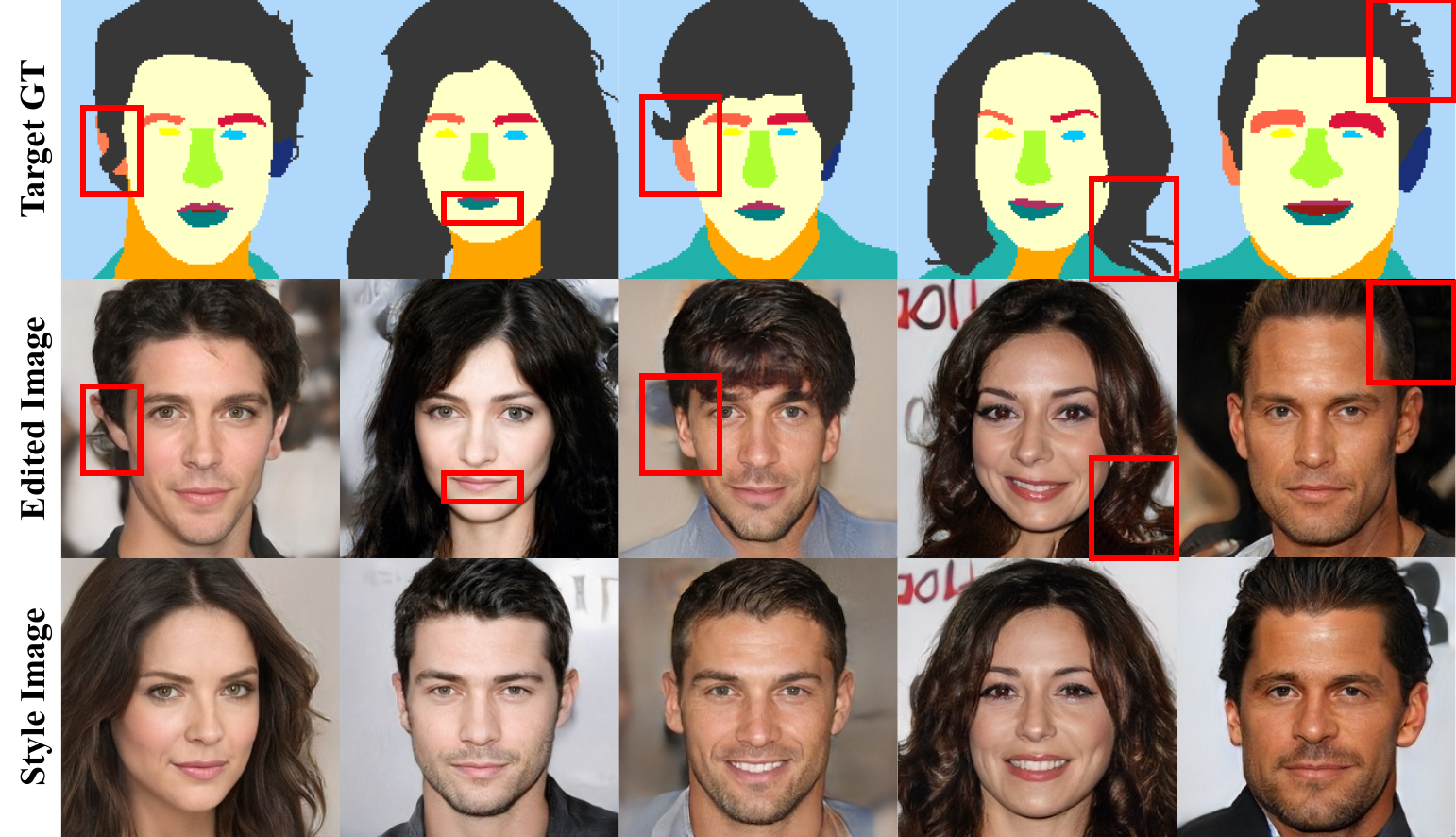}
  \caption{\textbf{Failure cases} of our method. Boxed areas show where the generated images do not match the corresponding masks.} 
  \label{fig:limitation}
\end{figure}

The proxy mask in our method is constructed from the clustering of intermediate feature maps of the pretrained Generator, especially size of 64. Consequently, the variance in resolution between the full image and the mask often results in an imperfect fit of the image to the given mask. \fref{fig:limitation} illustrates instances where our method encounters limitations due to this resolution difference.  In the highlighted red boxes within the figure, it is evident that intricate features, such as the length of a mouth or individual stray hairs, are not faithfully represented in the output image. This provides an opportunity for future work focused on enhancing pixel-level manipulability. 

\section{Conditional image synthesis using different types of conditions}

\begin{figure}[b!]
  \centering
  \begin{subfigure}[b]{\textwidth}
    \centering
    \includegraphics[width=\textwidth]{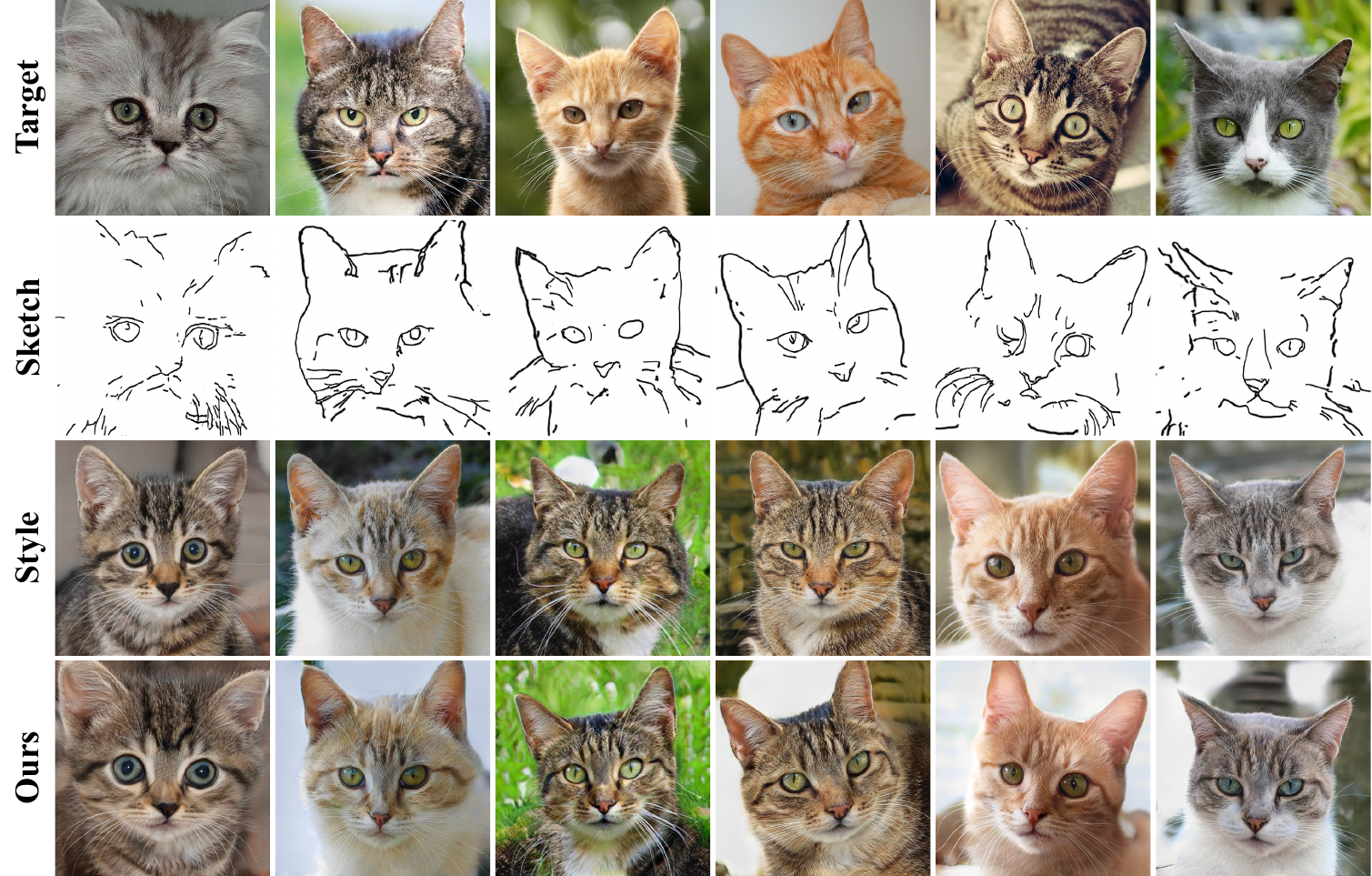}
    \caption{\textbf{Ours reflecting the sketch of cat}}
    \label{fig:sketch_cat}
    \vspace{8pt}
  \end{subfigure}
  \begin{subfigure}[b]{\textwidth}
    \centering
    \includegraphics[width=\textwidth]{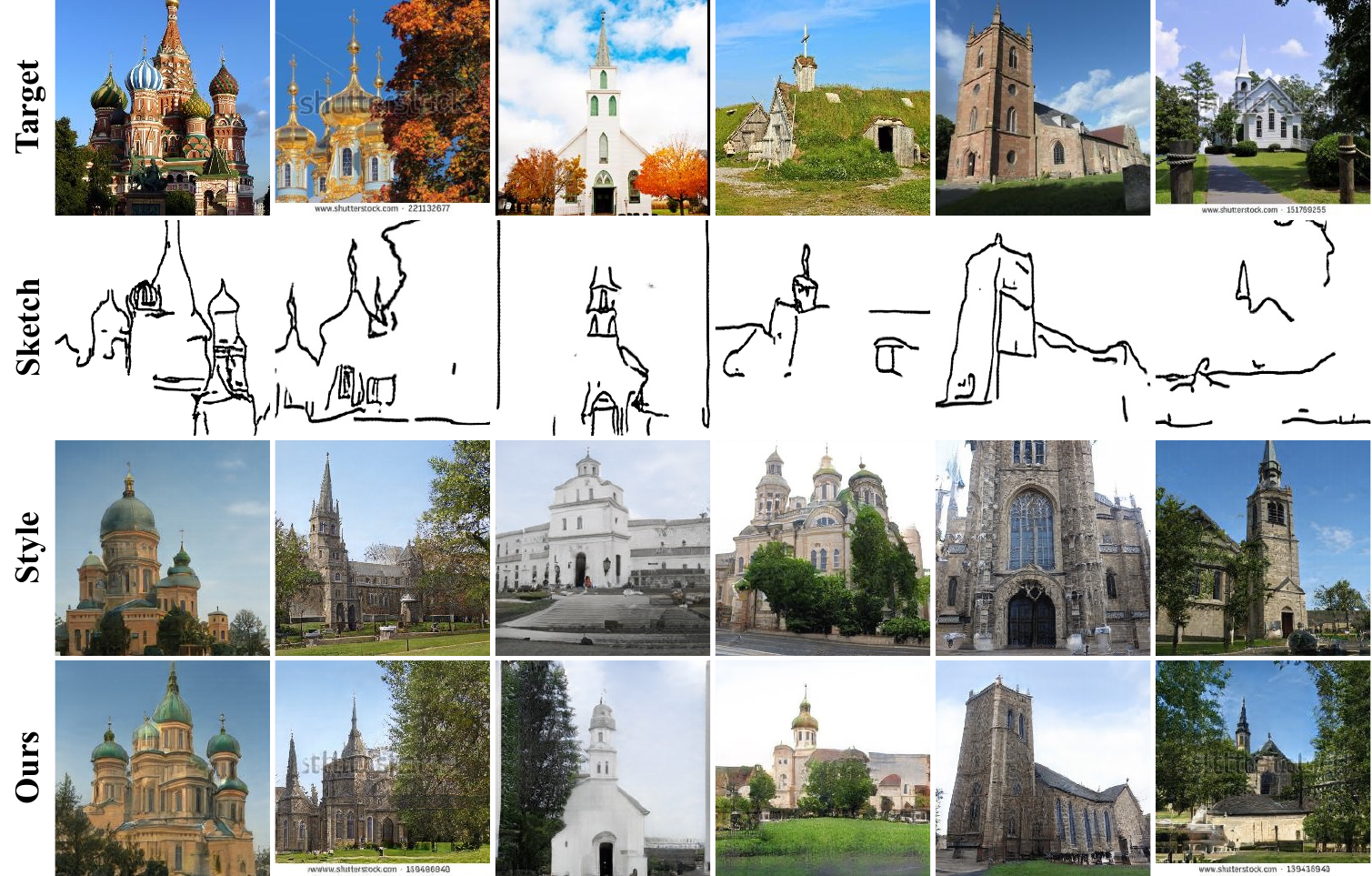}
    \caption{\textbf{Ours reflecting the sketch of church}}
    \label{fig:sketch_church}
  \end{subfigure}
  \caption{\textbf{Ours generated using the sketch on various dataset} (a) Ours accurately captures the shape obtained from the target image. The cat's facial angle in ours is appropriately aligned with the target shape. (b) Ours is capable of creating a church position similar to the church in the target image based solely on the sketch results. In addition, the style image's church, sky, and tree colors are applied equally in ours.}
  \label{fig:HEDsketch}
\end{figure}

\begin{figure}[h]
  \centering
  \includegraphics[width=\textwidth]{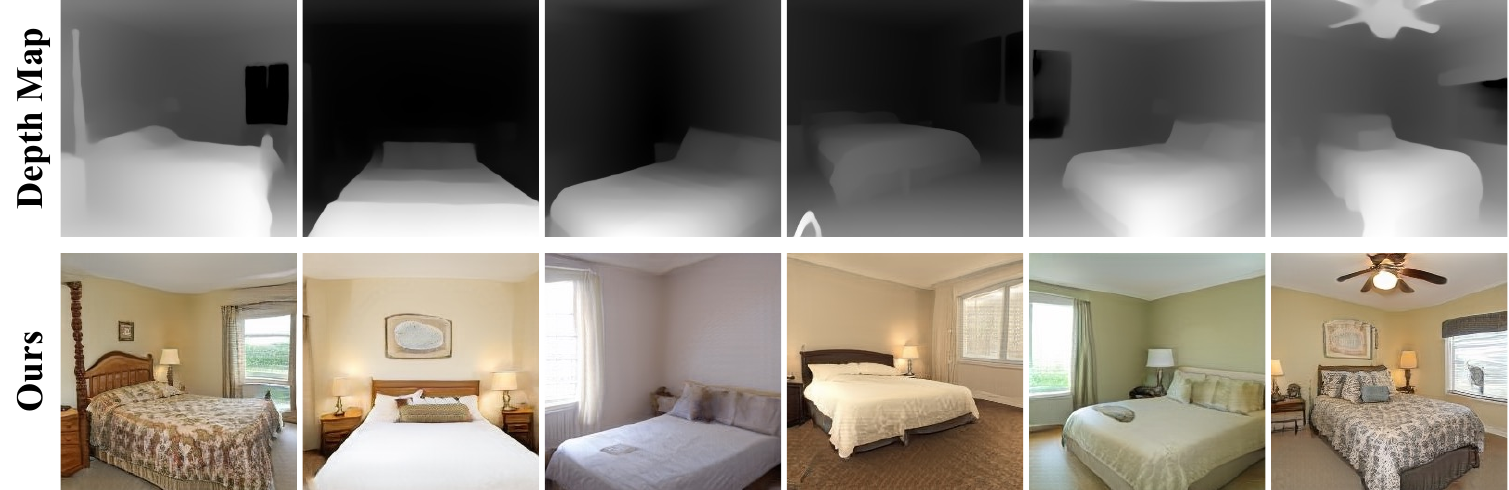}
  \caption{\textbf{Ours generated using the depth map on LSUN Bedroom} Ours effectively reflects the positions and shapes of objects such as beds or windows that can be observed in the depth map.}
  \label{fig:depth_bed}
\end{figure}

\begin{figure}[h]
  \centering
  \includegraphics[width=\textwidth]{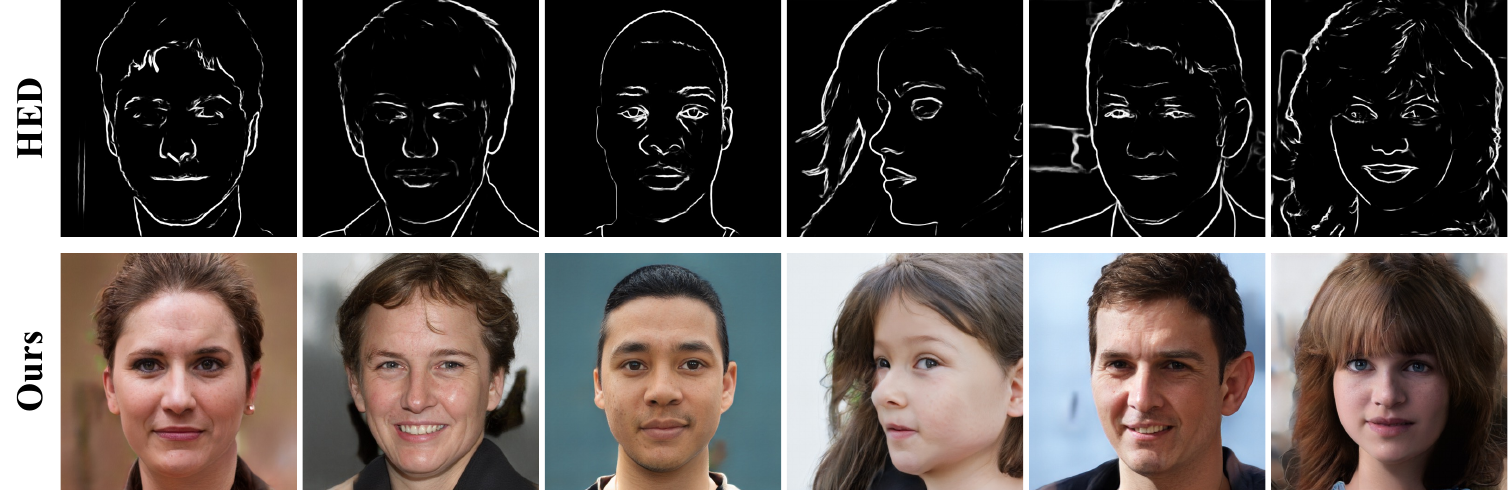}
  \caption{\textbf{Ours generated using the hed mask on FFHQ} Ours is generated to align with the facial shape determined by HED. It accurately captures aspects such as hairstyle, facial angle, and eye size in accordance with the shape outlined by HED.}
  \label{fig:depth_ffhq}
\end{figure}

In Figure \ref{fig:HEDsketch}, \ref{fig:depth_bed}, \ref{fig:depth_ffhq}, we present randomly sampled images generated by our model with different types of input conditions, utilizing Photo-sketch \cite{photo-sketching}, HED \cite{hed}, and depth \cite{midas} with various datasets such as FFHQ, AFHQ, LSUN Bedroom, and LSUN Church. We trained the mapper using conditions generated by passing training data through the corresponding pre-trained networks.

To generate training pairs for the mapper without relying on labor-intensive human annotations, we employed the methodology proposed by \cite{repurposegan} for generating segmentation maps and scribbles. Additionally, for the sketch, HED, Photo-sketch, and depth scenarios, we leveraged simple individual pretrained networks or tools to acquire pairs of generated images and their respective conditions.

% \subsection{Photo-sketch}
% \subsection{HED}
% \subsection{Depth}

\clearpage

\section{Semantic image synthesis with proxy mask}
\begin{figure}[t]
  \centering
  \includegraphics[width=\textwidth]{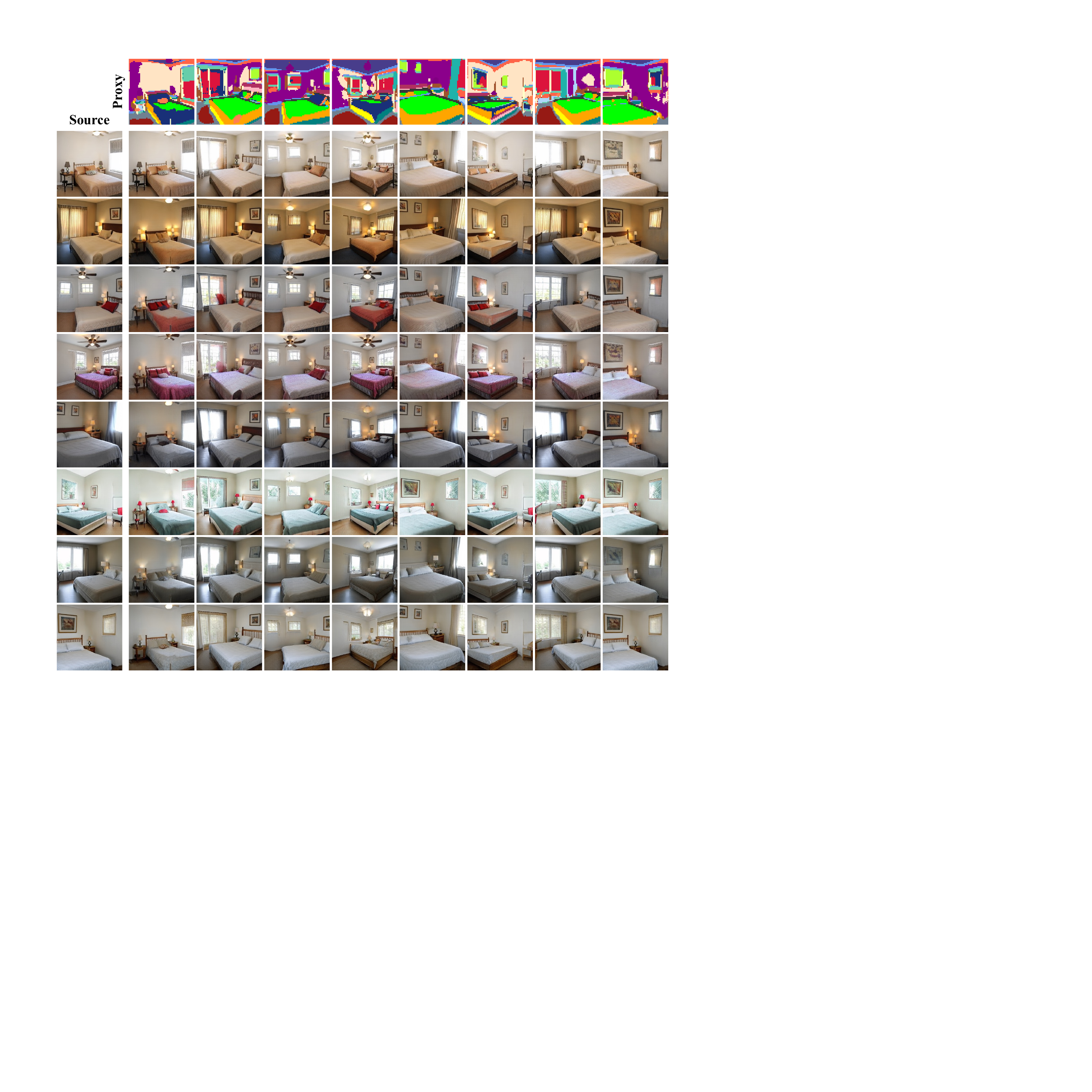}
  \caption{\textbf{Ours using the proxy mask obtained from random noise on LSUN Bedroom dataset} By generating images using the proxy mask without considering the perception gap, it is possible to create images of a higher quality.}
  \label{fig:Proxy_bed}
\end{figure}

\begin{figure}[h]
  \centering
  \includegraphics[width=\textwidth]{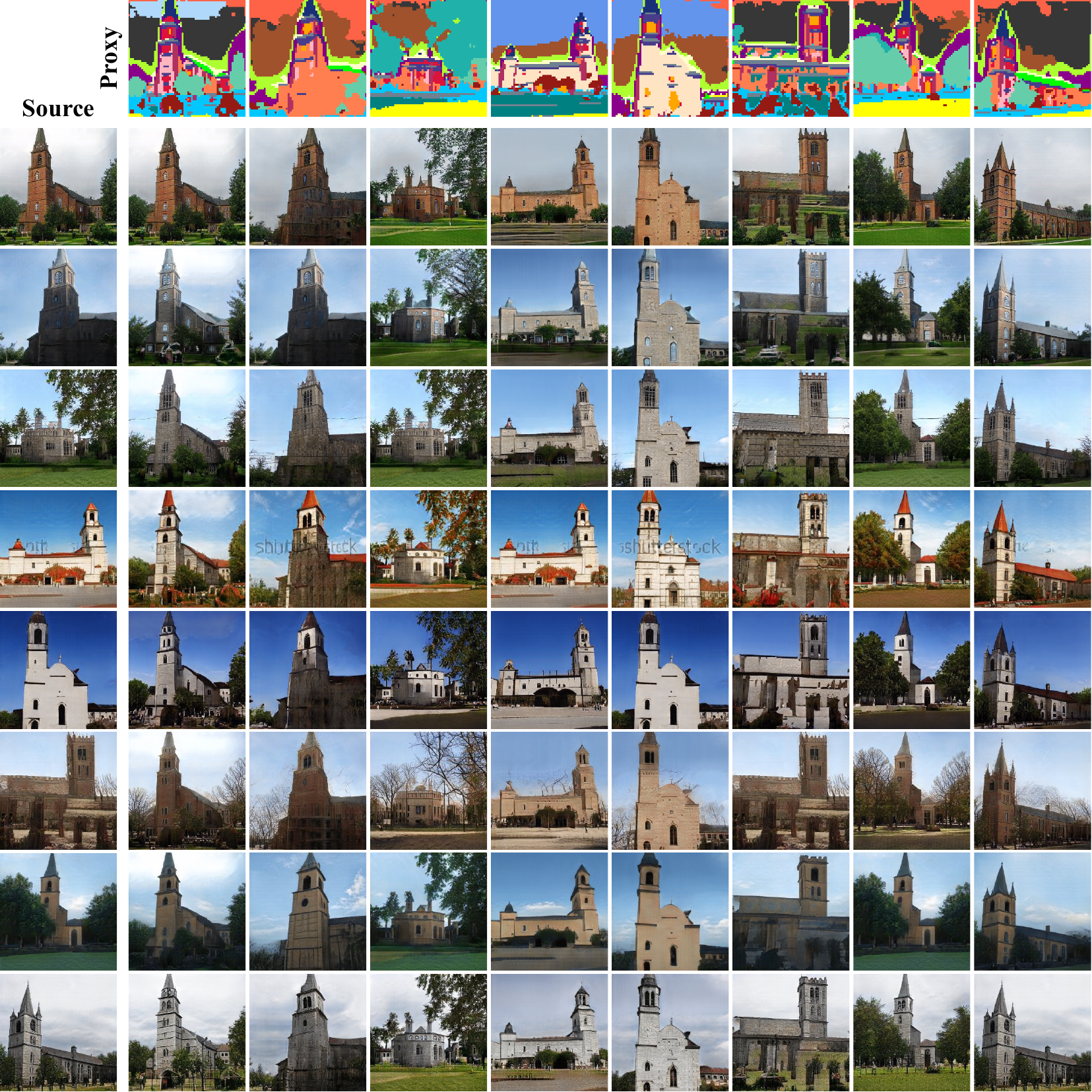}
  \caption{\textbf{Ours using the proxy mask obtained from random noise on LSUN Church dataset} By generating images using the proxy mask without considering the perception gap, it is possible to create images of a higher quality.}
  \label{fig:Proxy_church}
\end{figure}

\begin{figure}[h]
  \centering
  \includegraphics[width=\textwidth]{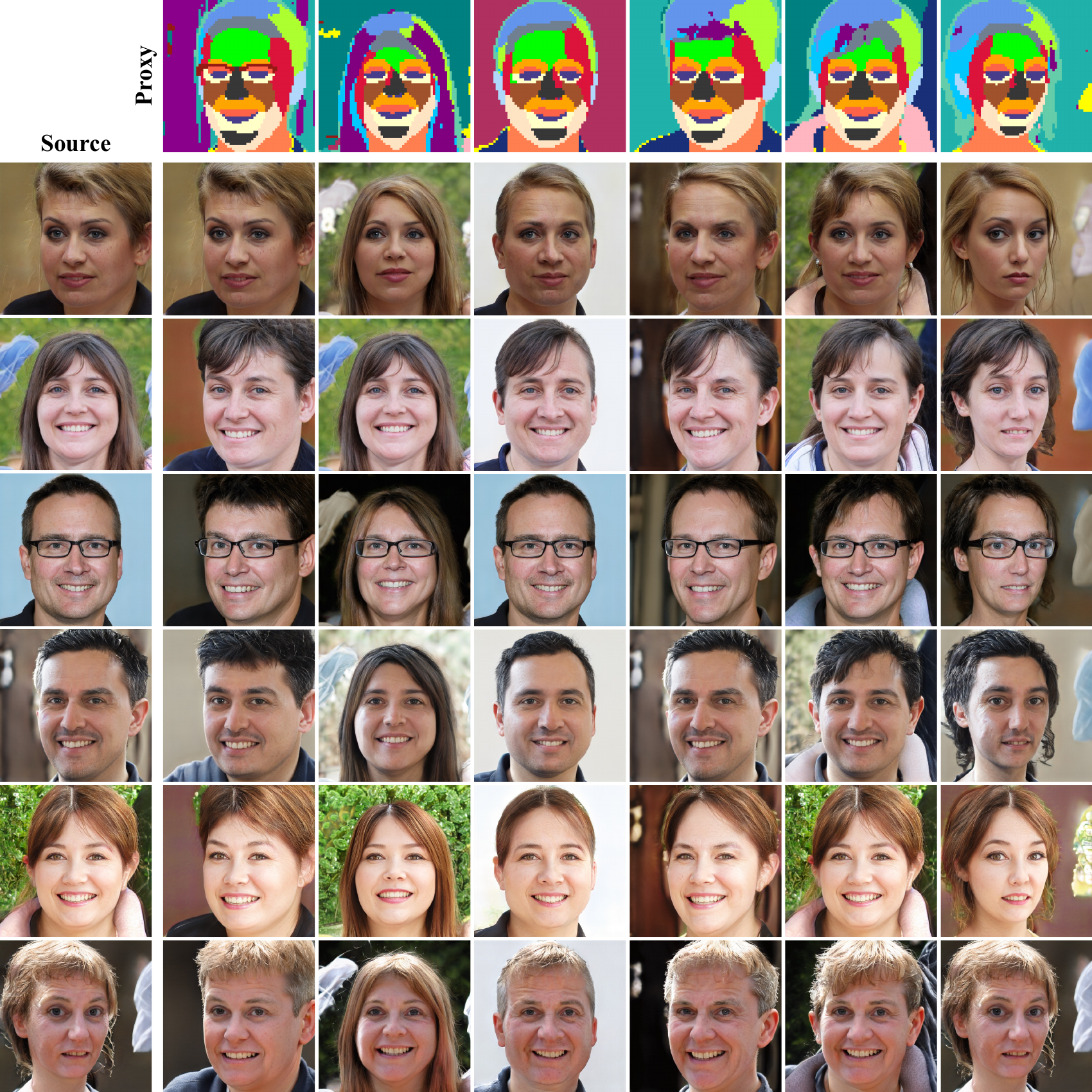}
  \caption{\textbf{Ours using the proxy mask obtained from random noise on FFHQ dataset} By generating images using the proxy mask without considering the perception gap, it is possible to create images of a higher quality.}
  \label{fig:Proxy_ffhq}
\end{figure}

\fref{fig:Proxy_bed}, \ref{fig:Proxy_church}, \ref{fig:Proxy_ffhq}, showcase the results of image generation solely using the proxy mask. Specifically, we generate images by conducting rearrangement the feature map with the proxy mask, both of which are derived from random noise. The output images illustrate the capability of our method to reflect both the attribute of the source and the shape of the target mask.


\clearpage
\section{More free-form image manipulation}
\begin{figure}[t!]
  \centering
  \begin{subfigure}[b]{0.49\textwidth}
    \centering
    \includegraphics[width=\textwidth]{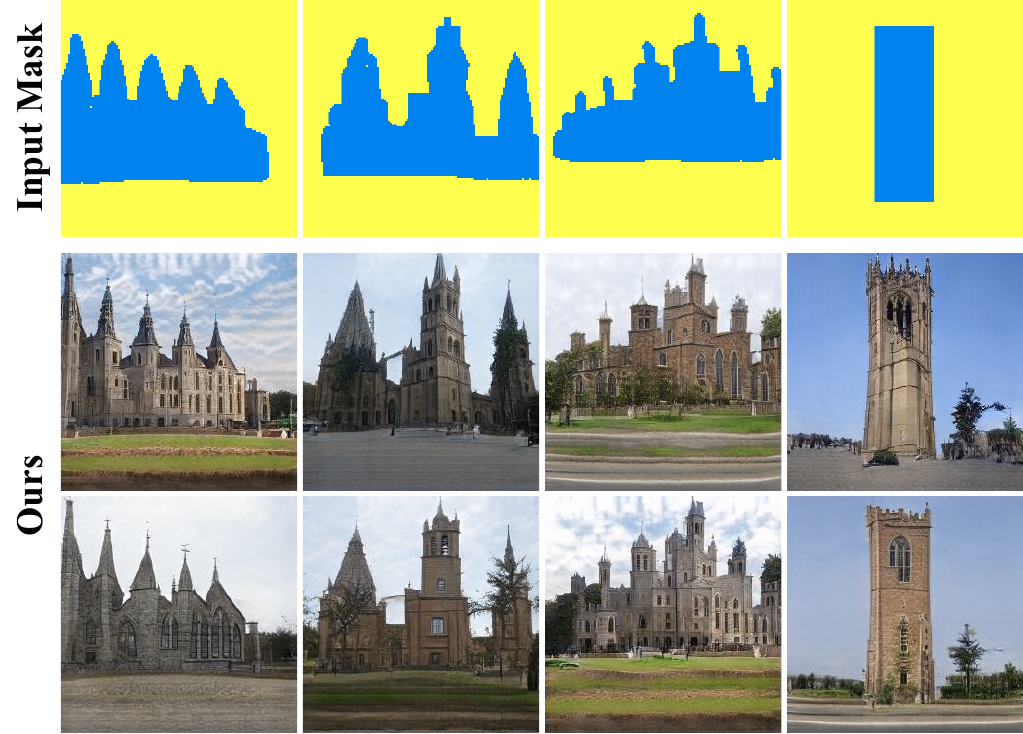}
    \caption{\textbf{Ours reflecting free-form mask on LSUN Church}}
    \label{fig:E_church}
  \end{subfigure}
  \begin{subfigure}[b]{0.49\textwidth}
    \centering
    \includegraphics[width=\textwidth]{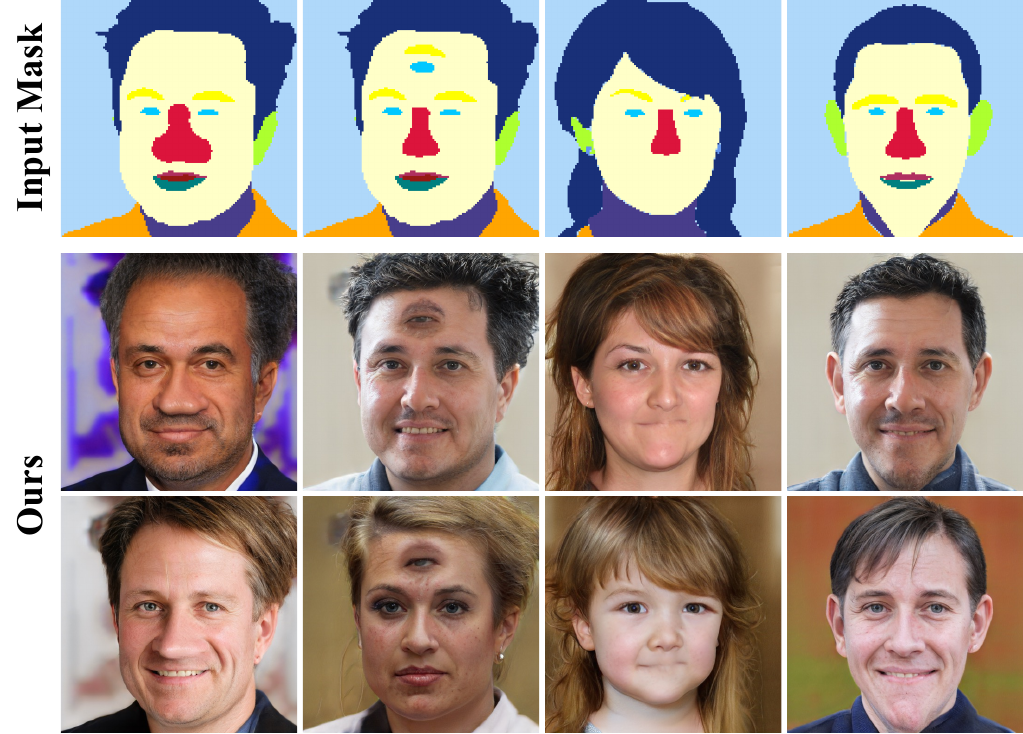}
    \caption{\textbf{Ours reflecting free-form mask on FFHQ}}
    \label{fig:E_ffhq}
  \end{subfigure}
  \caption{\textbf{Free-form image manipulation} When given free-form mask images, our model flexibly synthesizes images. Even masks with unusual shape, such as a rectangular mask of LSUN Church or a three-eyed mask for FFHQ, can generate proper images.}
  \label{fig:E_freeform}
\end{figure}

 In \fref{fig:E_freeform}. We show the results using free-form image manipulation on LSUN Church and FFHQ.

% {\small
% \bibliographystyle{ieee_fullname}
% \bibliography{egbib_sup}
% }

\end{appendix}

% \end{document}

% \cho{Ours CelebA Freeform 보여주어야할까 - 굳이 보여줄 필요은 없다고 생각하지만 본문에서 OASIS는 celeba인데 ours는 ffhq면 안 맞는 설정이 아니냐는 질문이 들어올 수도 있을듯함. CelebA에서 해보고 잘 되어야 넣을 수 있을듯함. / 다른 마스크 형태에 대해서 free-form 보여주기(본문에 넣기는 애매하지만 좀 더 이것까지 되네라는걸 넣어주면 좋을듯함 ex: 스펀지밥) / Church나 고양이 등 다양한 데이터에서 본문에 보여주진 않았지만 OOD shape에 대해서도 이미지들을 생성해낼 수 있음을 보여주기 - SPADE, OASIS도 같이 보여주면 좋을텐데 보여줄 수 없으니 적당히 ADE20K? / 같은 섹션에 보여주어야할지는 고민이 되지만 조금 더 다양한 케이스의 어노테이션(ex: label 하나씩 늘리기)에 대해서도 빠른 학습으로 이미지를 보여줄 수 있음을 보여주면 좋을듯함.}

% \section{Unaligned local transplantation}
% % 가능하면 넣기
% \cho{real image 사진들 뽑아둔 걸로 진행하면 될듯(잘 되면 넣고 아니면 아예 삭제가 나을듯해요)}

% \section{Exemplar 내용}
% \cho{더 넣어야할까요? real끼리 하면 TransEditor가 예상보다 disentangle이 잘 되어서 애매하네요}

% {\small
%     \bibliography{egbib}
%     \bibliographystyle{plainnat}
% }

\end{document}